\pdfoutput=1
\documentclass[10pt,twocolumn,letterpaper]{article}

\usepackage[pagenumbers]{cvpr}

\usepackage{booktabs}
\usepackage{amsmath,amssymb,amsthm}
\newtheorem{proposition}{Proposition}
\usepackage{tikz}
\usetikzlibrary{arrows.meta,positioning}

\newcommand{\robotw}{0.9}

\definecolor{cvprblue}{rgb}{0.21,0.49,0.74}
\usepackage[pagebackref,breaklinks,colorlinks,allcolors=cvprblue]{hyperref}

\title{Planning as Likelihood Evaluation: A Score-Based Theory of Joint-Embedding World Models}

\author{Junyi Hu \qquad Zhewen He \qquad Shuaihang Yuan \qquad Haomian Huang\\
Jiazhao Liang \qquad Zhenhua Li \qquad Yi Fang\thanks{Corresponding author.}\\[4pt]
New York University Abu Dhabi
}

\begin{document}
\maketitle
\begin{abstract}
Joint-embedding world models predict the future in representation space, and a robot plans with them by scoring imagined outcomes against a goal. The score is written by hand: a distance between the latent rolled out to the horizon and the goal latent. It treats an implausible goal like a likely one, sees the goal at one fixed scale, accepts only image goals, and costs a rollout per candidate. We take a different view of the same model. Under denoising score matching, the predictor's objective at a goal bounds the negative log-likelihood of that goal under the candidate actions, and the hand-written distance is this bound's term at the highest noise level. Planning is then likelihood evaluation: one forward pass per candidate, with no rollout and no distance function. The noise level sets the resolution of the cost, the training horizon sets its range, and goals multiply, so an image, several views and a language instruction condition one energy. The same reading says how to train the predictor: predict the clean target at the noise levels where the actions are read. The resulting model, S-JEPA, plans CALVIN manipulation through a frozen video encoder from the instruction alone, completing 63\% of the validation windows against 31\% for the hand-written distance from a goal image. Where the actions leave the future undetermined, it reaches the goal in 0.86 of instances against 0.38. Read from the released navigation world model without training, the energy reaches the official rollout planner's error in a fortieth of its time, and on PointMaze it matches the DINO-WM planner in a sixteenth. On a real robot arm it completes 85 of 100 trials from a goal image against 60 for the hand-written distance, and 94 from typed instructions.
\end{abstract}

\section{Introduction}
\label{sec:intro}

\begin{figure}[t]
\centering
\input{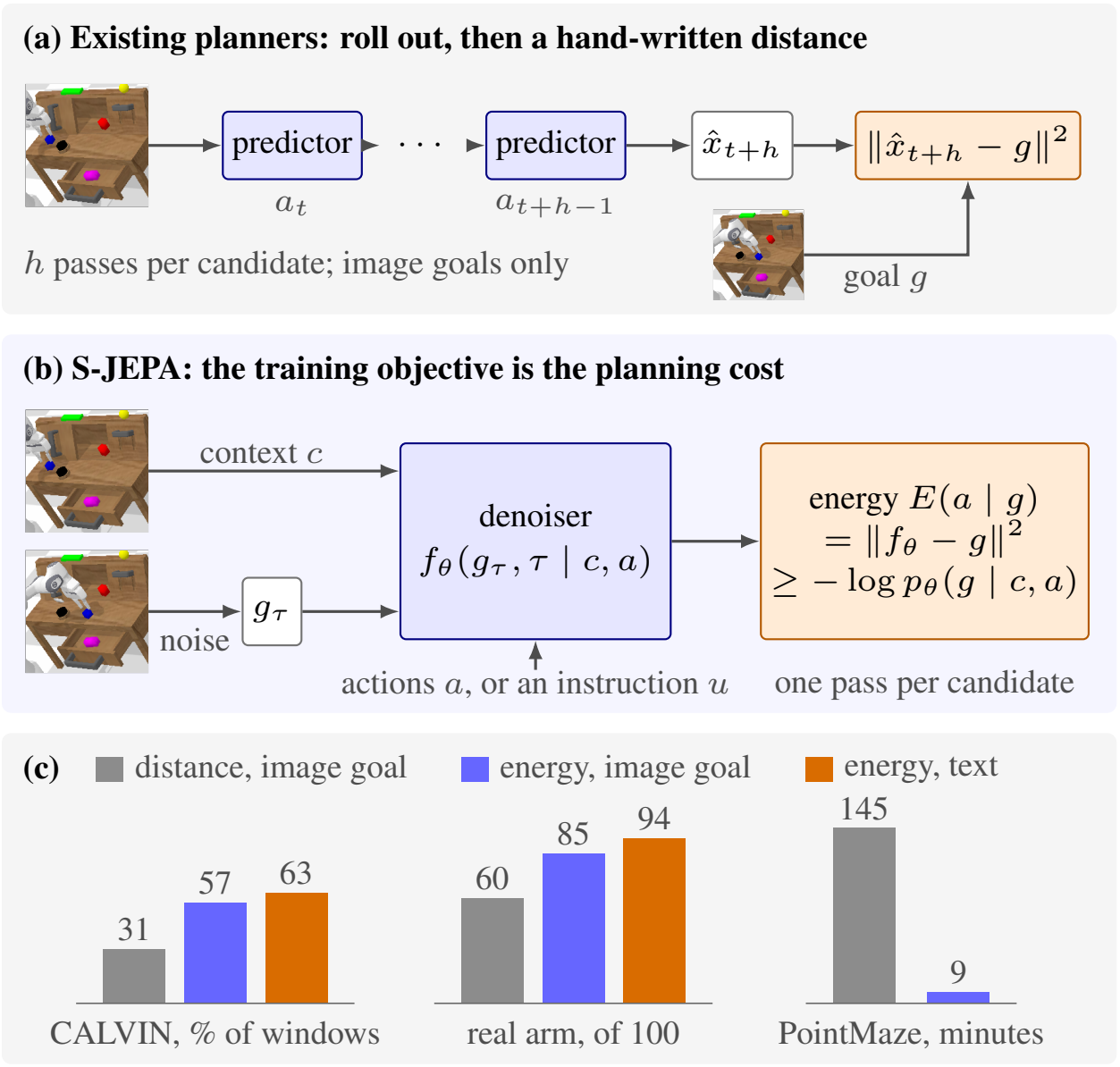}
\caption{Planning as likelihood evaluation. (a) Existing planners roll the predictor out along each candidate and score the endpoint with a distance written after training. (b) S-JEPA reads the cost from the predictor's own denoising objective at the noised goal $g_\tau$: one forward pass bounds the negative log-likelihood of the goal under the actions. (c) Trials completed with the same predictor architecture, data and budget, and planning time on PointMaze.}
\label{fig:teaser}
\end{figure}

World models that predict in representation space have become the way embodied agents imagine the consequences of their actions~\cite{lecun2022path,ijepa,bardes2024vjepa,dinowm,vjepa2,vjepa21,nwm,adwm}. A joint-embedding predictor is trained once, on video and the actions taken in it, and the promise is that the same predictor then plans for any goal: a target image, a view of the scene, a demonstration, eventually a sentence. In practice the prediction is only half of planning. The other half is a cost that says how good an imagined future is, and in every joint-embedding world model deployed so far that cost is written by hand after training~\cite{dinowm,vjepa2,nwm}. The predictor is rolled out along a candidate action sequence, the final latent is compared with the latent of a goal image by a distance, and the cross-entropy method searches for the sequence with the smallest distance. The distance does not know which coordinates of the representation matter for the task, it treats a goal the model finds implausible exactly like one it finds likely, and it examines the goal at one fixed scale, layout and detail alike. It can only be taken to an image, and every candidate costs a rollout to the horizon, so the search runs for minutes per decision. The model is trained to predict and asked to plan, and the two are joined by a cost that nobody trained.

The remedies in use keep the cost where it is. Regularising the predictor so that its latents retain action information~\cite{adwm} sharpens the distance but does not say which goals are reachable; a learned value function or a vision-language reward~\cite{liv,vlmrm} adds a second model whose errors the planner cannot see; and language goals are reached by generating a goal image or a video~\cite{susie,unipi} that is then handed to the same distance, or by giving up planning for a policy~\cite{mdt,gr1,3dda}. None asks what the predictor knows about a goal: a deterministic predictor knows only the mean of the future, and a distance to that mean is all it can answer.

We take a different view of the same model (Fig.~\ref{fig:teaser}). A predictor trained by denoising score matching learns, at every noise level, how to move a noised target toward the futures that the context and the actions allow. By the maximum-likelihood reading of score matching~\cite{song2021maximum,kingma2023understanding}, its training loss evaluated at a goal, averaged over noise levels and noise draws, is a bound on the negative log-likelihood of that goal under the actions. Planning is then likelihood evaluation: one forward pass per candidate and noise draw, no rollout and no distance function. The view contains the hand-written cost. At the highest noise level the noised goal carries no information, the denoiser returns the conditional mean of the future, and the objective reduces to the latent distance between prediction and goal; latent-distance planning is the high-noise limit of score-based planning. The view also answers three questions that the distance cannot ask. The noise level is the resolution at which the cost sees the goal, from the far field that guides a search down to the details of a goal already in view. The training horizon is the range over which the cost carries information about the goal, and it is extended by training longer, with the same planner. And a likelihood multiplies: goal images, views of a goal and the end states of demonstrations enter one energy as a product of experts, and a language instruction enters as a goal sampled from the predictor's own conditional. The same reading says how the predictor must be trained if its objective is to serve as the cost: the actions are read only at the noise levels where the noised goal does not reveal itself, so the predictor should predict the clean target and train mostly at high noise. We call this predictor S-JEPA, a joint-embedding world model whose objective is its planning cost; it also carries the instruction in its action slot and trains its encoder against a moving-average target.

Our contributions are: (i)~a view of planning in joint-embedding world models in which the training objective is the cost, stated in three propositions, namely that the energy is a conditional-likelihood bound, that the latent distance is its high-noise limit, and that the conditional channel is trained only where the noised target does not reveal itself; (ii)~S-JEPA, the predictor these propositions prescribe, with the measurements on PushT and CALVIN that tie the parametrisation, the noise schedule and the horizon of a predictor to the cost that can be read from it; and (iii)~demonstrations in which one energy plans an arm from images or from language alone, in CALVIN and on the real robot, the released navigation world model without training, the DINO-WM benchmarks at a small fraction of the rollout planner's time, and a stochastic environment in which it ranks actions by the probability of reaching the goal.

\section{Related Work}
\label{sec:related}

JEPA~\cite{lecun2022path,ijepa,bardes2024vjepa} predicts in representation space, and the planners built on it share one recipe: DINO-WM~\cite{dinowm}, V-JEPA~2-AC~\cite{vjepa2,vjepa21} and NWM~\cite{nwm} roll the predictor to the horizon for every candidate and rank the candidates by a distance to the goal embedding, which AD-WM~\cite{adwm} sharpens by regularising the latents. Language reaches these planners from outside: a generated goal image or video~\cite{susie,unipi}, a learned reward~\cite{liv,vlmrm}, or a policy trained on annotated play~\cite{calvin,mdt,gr1,3dda}. LeCun~\cite{lecun2022path} frames the architecture as an energy-based model and leaves the planning cost to a separate module. We show that for a predictor trained by denoising score matching the training energy is that cost, and the instruction conditions the same predictor.

That objective is denoising score matching across noise levels~\cite{vincent2011connection,song2019generative,song2020improved,ho2020ddpm,song2021sde}. Song et al.~\cite{song2021maximum} relate its weighted form to a likelihood bound, Kingma and Gao~\cite{kingma2023understanding} characterise the weightings that keep this reading, and P2 and EDM~\cite{choi2022p2,karras2022elucidating} locate the levels at which the denoiser learns content; conditional scores compose by addition~\cite{liu2022composable,du2023reduce}. The objective has been read as a likelihood to classify images~\cite{li2023diffusionclassifier} and to reward reinforcement learning from a video model of expert data~\cite{escontrela2023viper,huang2024diffusionreward}; in control, diffusion otherwise generates trajectories~\cite{janner2022diffuser,chi2023diffusion}, and model-based agents score rolled futures with a separate reward~\cite{hafner2023dreamerv3,hansen2024tdmpc2}. The model whose likelihood we read is conditioned on actions, so the likelihood ranks actions instead of classes, no expert data defines a reward, and the planner never samples the future it scores.

\section{Planning as Likelihood Evaluation}
\label{sec:theory}

\subsection{Setting}
\label{sec:setting}

An encoder maps a frame $o_t$ to $P$ tokens $z_t \in \mathbb{R}^{P\times D}$. We write $x$ for the target representation of a frame: its tokens and one token holding the proprioceptive state, standardised per dimension. A joint-embedding world model has a predictor that, given a context $c = (z_{t-k+1:t}, a_{t-k+1:t})$ of $k$ frames with the actions taken at them and a chunk $a = a_{t:t+h-1}$ of $h$ future actions, describes the representation $x_{t+h}$ of the frame $h$ steps ahead. A deterministic predictor regresses it, $\hat{x}(c, a) \approx \mathbb{E}[x_{t+h} \mid c, a]$, with an $L_1$ or $L_2$ loss~\cite{dinowm,vjepa2}. Planning toward a goal $g$ (the target representation of a goal frame) then minimises a weighted latent distance over candidate chunks,
\begin{equation}
C_{\mathrm{dist}}(a \mid g) = \sum_i w_i \,\lVert \hat{x}_i(c, a) - g_i \rVert^2 ,
\label{eq:dist}
\end{equation}
with weights $w_i$ uniform or set from demonstrations.
\label{sec:gaussian}Up to a constant, Eq.~\ref{eq:dist} is the negative log-density of a diagonal Gaussian with mean $\hat{x}(c,a)$ and precisions $w_i$: the planner scores a chunk by how likely the goal is under a Gaussian whose spread is fixed by hand. It has one mode, so when the actions allow several futures the planner is pulled toward a latent that no future occupies, and its weights depend neither on the task nor on the scale at which the goal is examined. A score model learned on the same latents removes these restrictions and contains the Gaussian as a limit.

\subsection{The objective at a goal is a likelihood bound}
\label{sec:bound}

\begin{figure}[t]
\centering
\includegraphics[width=0.9\linewidth]{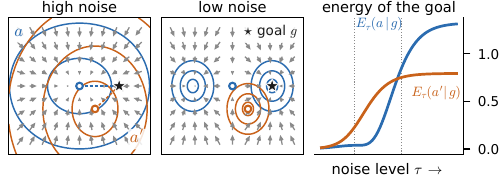}
\caption{The energy reads a likelihood from the conditional score field (closed-form example). Action $a$ (blue) allows two futures, one of them the goal $g$; $a'$ (orange) allows one, closer to $g$ than the mean of $a$. Arrows: $\nabla \log p_\tau(\cdot \mid c, a)$. At high noise the energy is the distance to each mean (dashed) and $a'$ is cheaper; at low noise the futures of $a$ separate, its score points back to $g$, and $a$ becomes cheaper (right; dotted: the two levels shown).}
\label{fig:field}
\end{figure}

Let the predictor be a conditional denoiser $f_\theta(x_\tau, \tau \mid c, a)$ trained by denoising score matching on the target representation. With a variance-preserving schedule $\bar\alpha_\tau$, $\tau \in \{1,\dots,T\}$, a noised target is $x_\tau = \sqrt{\bar\alpha_\tau}\, x + \sqrt{1-\bar\alpha_\tau}\,\epsilon$, and the $x_0$-parametrised objective is
\begin{equation}
\mathcal{L}(\theta) = \mathbb{E}_{(c,a,x)}\, \mathbb{E}_{\tau\sim p(\tau),\,\epsilon}\;
\big\lVert f_\theta(x_\tau, \tau \mid c, a) - x \big\rVert^2 .
\label{eq:dsm}
\end{equation}
The denoiser is a model of the conditional score of the noised future~\cite{vincent2011connection,song2021sde},
\begin{equation}
\begin{aligned}
s_\theta(x_\tau, \tau \mid c, a) &= \frac{\sqrt{\bar\alpha_\tau}\, f_\theta(x_\tau, \tau \mid c, a) - x_\tau}{1-\bar\alpha_\tau}\\
&\approx \nabla_{x_\tau} \log p_\tau(x_\tau \mid c, a),
\end{aligned}
\label{eq:score}
\end{equation}
and Eq.~\ref{eq:dsm} is denoising score matching for $s_\theta$ with weight $(1-\bar\alpha_\tau)^2/\bar\alpha_\tau$ on level $\tau$~\cite{ho2020ddpm}. We define the energy of a chunk at a goal as the same objective with the goal in place of the sampled future,
\begin{equation}
\begin{aligned}
E(a \mid g) &= \sum_{\tau \in \mathcal{T}} \lambda_\tau\, \mathbb{E}_{\epsilon}\;
\big\lVert f_\theta(g_\tau, \tau \mid c, a) - g \big\rVert^2 ,\\
g_\tau &= \sqrt{\bar\alpha_\tau}\, g + \sqrt{1-\bar\alpha_\tau}\,\epsilon ,
\end{aligned}
\label{eq:energy}
\end{equation}
with levels $\mathcal{T}$, weights $\lambda_\tau$, and $g_\tau$ the goal noised to level $\tau$. Substituting Eq.~\ref{eq:score} writes the term $E_\tau$ of each level as a score-matching residual at the noised goal,
\begin{equation}
E_\tau(a \mid g) = s_\tau\, \mathbb{E}_{\epsilon}\, \big\lVert \epsilon + \sqrt{1-\bar\alpha_\tau}\; s_\theta(g_\tau, \tau \mid c, a) \big\rVert^2 ,
\label{eq:residual}
\end{equation}
with $s_\tau = (1-\bar\alpha_\tau)/\bar\alpha_\tau$ the noise-to-signal ratio. The energy is low when the score field of the futures that the actions allow points from the noised goal back to the goal, that is, when it explains the noise that was added (Fig.~\ref{fig:field}).

\begin{proposition}[Likelihood bound]
\label{prop:bound}
Let $p_\theta(\cdot \mid c, a)$ be the density that the score model of Eq.~\ref{eq:score} defines through its reverse diffusion, and $\gamma = 1/s_\tau$ the signal-to-noise ratio. In the continuous-time limit of the schedule,
\begin{equation}
-\log p_\theta(g \mid c, a) \;\le\; \frac12 \int_0^\infty\! E_{\tau(\gamma)}(a \mid g)\, d\gamma \,+\, C ,
\label{eq:bound}
\end{equation}
with $C$ independent of $a$ and equality when the score model is exact~\cite{song2021maximum}. The right-hand side is the energy $E^\star$ of Eq.~\ref{eq:energy} over all levels with the likelihood weighting $\lambda^\star_\tau = \tfrac12(\gamma_{\tau-1} - \gamma_\tau)$. For any weighting monotone in the noise level, $E(a \mid g)$ is, up to a constant, the negative evidence lower bound of $g$ under $p_\theta(\cdot \mid c, a)$ with Gaussian noise added to the data at the levels it emphasises~\cite{kingma2023understanding}.
\end{proposition}

The proposition restates two published results with the conditioning made explicit (Appendix~\ref{app:proofs}). Minimising $E$ over chunks maximises a likelihood bound of the goal under the actions, so one forward pass per candidate replaces a rollout. Restricting the integral to the levels $\tau \ge \tau_{\min}$ bounds the likelihood of the goal noised to $\tau_{\min}$, a goal specified to within $\sqrt{s_{\tau_{\min}}}$ per standardised dimension. The planner reads this bound at a few levels above $\tau_{\min}$; by the next two propositions, the level is a resolution.

\begin{figure*}[t]
\centering
\begin{tikzpicture}[x=1mm, y=1mm,
  box/.style={draw, rounded corners=1.2pt, minimum height=5mm, inner sep=2pt, align=center, font=\scriptsize},
  lat/.style={box, fill=white, draw=black!55},
  enc/.style={box, fill=black!8, draw=black!50},
  net/.style={box, fill=blue!10, draw=blue!50!black},
  cost/.style={box, fill=orange!16, draw=orange!70!black},
  tok/.style={draw=black!55, fill=white, minimum width=2.2mm, minimum height=2.2mm, inner sep=0pt},
  note/.style={font=\scriptsize, text=black!70, align=left, inner sep=0pt},
  ttl/.style={font=\scriptsize\bfseries, anchor=west, inner sep=0pt},
  arr/.style={-{Latex[length=1.4mm]}, semithick, black!70},
]
\fill[black!4, rounded corners=2pt] (0,0) rectangle (60,-38);
\node[ttl] at (1.5,-2.7) {(a) Training: denoise the future representation};
\foreach \i in {0,1,2,3} \node[tok, fill=black!12] at (4.5+2.7*\i,-30.5) {};
\foreach \i in {0,1,2} \node[tok, fill=blue!22] at (24.3+2.7*\i,-30.5) {};
\foreach \i in {0,1,2,3} \node[tok, fill=orange!30] at (42+2.7*\i,-30.5) {};
\node[note] at (8.5,-34.7) {context $c$}; \node[note] at (27,-34.7) {slot: $a$ or $u$}; \node[note] at (46,-34.7) {noised target $x_\tau$};
\node[net, minimum width=47mm, minimum height=8.5mm] (f) at (27,-20.5) {predictor $f_\theta$: one transformer,\\block mask, embeddings of $\tau$ and $h$};
\draw[arr] (8.5,-28.9) -- (8.5,-24.9); \draw[arr] (27,-28.9) -- (27,-24.9); \draw[arr] (46,-28.9) -- (46,-24.9);
\node[cost] (L) at (20,-9.3) {$\lVert f_\theta(x_\tau,\tau\mid c,a) - x \rVert^2$};
\draw[arr] (f.north -| L) -- (L);
\node[note, anchor=west] (xt) at (41.5,-7.6) {clean target $x$};
\node[note, anchor=west] at (41.5,-10.8) {(EMA encoder)};
\draw[arr] (40.9,-9.3) -- (L.east);
\node[note, anchor=west] at (38,-14.6) {$\tau$: half in $[0.7T,T)$};
\fill[black!4, rounded corners=2pt] (62,0) rectangle (112,-38);
\node[ttl] at (63.5,-2.7) {(b) Goals from an instruction};
\node[lat, font=\scriptsize\itshape] (u) at (87,-9.3) {``open the drawer''};
\node[enc] (clip) at (87,-17.3) {CLIP text, one token $u$ in the slot};
\node[net] (fb) at (87,-25.3) {$f_\theta$, DDIM, 10 steps};
\node[lat] (gk) at (87,-33.3) {$g^{(1)},\dots,g^{(K)} \sim p_\theta(\cdot \mid c,u)$};
\draw[arr] (u) -- (clip); \draw[arr] (clip) -- (fb); \draw[arr] (fb) -- (gk);
\fill[blue!4, rounded corners=2pt] (114,0) rectangle (174.6,-38);
\node[ttl] at (115.5,-2.7) {(c) Planning: evaluate the goal, do not roll out};
\node[lat] (gt) at (127.3,-10.3) {goals $g_m$, noised\\to $g_{m,\tau}$ (shared $\epsilon$)};
\node[lat] (cem) at (127.3,-22) {$N$ chunks\\$a^{n}\!\sim\!\mathcal{N}(\mu,\sigma^2)$};
\node[net, minimum height=19.5mm] (fc) at (146.3,-16.15) {$f_\theta$\\[0.3mm]one pass\\per chunk};
\node[cost, minimum height=19.5mm] (E) at (163.9,-16.15) {$\sum_m E(a^{n}\!\mid g_m)$\\[0.5mm]$\ge -\!\sum_m \log$\\$p_\theta(g_m\!\mid c,a^{n})$};
\draw[arr] (cem.east) -- (cem.east -| fc.west); \draw[arr] (gt.east) -- (gt.east -| fc.west); \draw[arr] (fc) -- (E);
\draw[arr] (E.south) |- (146,-29.2) -| (cem.south);
\node[note] at (145.5,-31.4) {refit $\mu,\sigma$ to the $M$ lowest energies, $R$ rounds};
\node[note, anchor=west] at (115.5,-35.6) {then execute $e$ actions; stop when $E(\text{do nothing})$ flattens};
\end{tikzpicture}
\caption{S-JEPA. (a) One transformer denoises the target representation of the frame $h$ steps ahead given the context and the action slot; it predicts the clean target and draws half of its noise levels from the upper third of the schedule. (b) An instruction takes the action slot, and the same network samples goal latents. (c) The planner searches chunks by the cross-entropy method on the energy: every candidate costs one forward pass at the noised goals, and goals from images, views or an instruction add.}
\label{fig:pipeline}
\end{figure*}
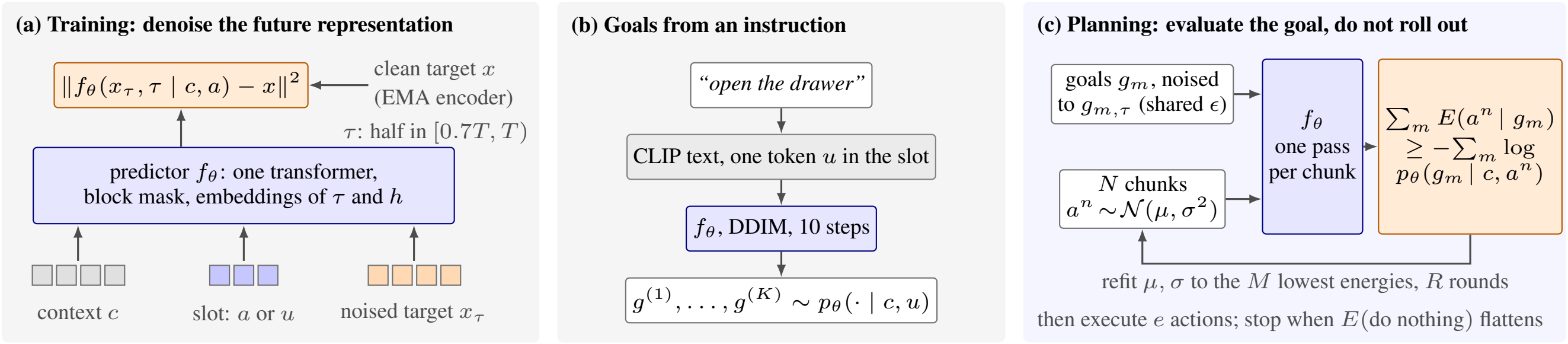

\subsection{The latent distance is the high-noise limit}
\label{sec:limit}

\begin{proposition}[High-noise limit]
\label{prop:limit}
As $\bar\alpha_\tau \to 0$ the noised goal carries no information about $g$, the optimal denoiser is the conditional mean, $f_\theta(x_\tau, \tau \mid c, a) \to \mathbb{E}[x \mid c, a]$, and the term of the energy at that level, $E_\tau(a \mid g) = \mathbb{E}_{\epsilon} \lVert f_\theta(g_\tau, \tau \mid c, a) - g \rVert^2$, is the squared latent distance,
\begin{equation}
E_\tau(a \mid g) \;\longrightarrow\; \big\lVert \mathbb{E}[x \mid c, a] - g \big\rVert^2 .
\label{eq:limit}
\end{equation}
\end{proposition}

Eq.~\ref{eq:limit} is Eq.~\ref{eq:dist} with uniform weights: the latent-distance planner of DINO-WM and V-JEPA~2-AC is the energy at its highest noise level, with the deterministic predictor standing in for the conditional mean. Everything the Gaussian view lacks sits at the lower levels, where the noised goal shows first its layout and then its details and the denoiser has to reconcile them with the futures the actions allow: the energy at a level is a likelihood of the goal at that resolution. When a chunk allows a single Gaussian future, every level orders the chunks as the limit does (Appendix~\ref{app:proofs}); the energy departs from the distance where the actions leave the future undetermined (Sec.~\ref{sec:stoch}).

\subsection{Which noise levels carry the condition}
\label{sec:channel}

Not every level is useful for ranking actions. Let the actions allow $K$ futures, $x \sim \sum_k \pi_k\, \mathcal{N}(\mu_k, \sigma^2 I)$ with $\pi_k$ and $\mu_k$ set by the chunk. In $z_\tau = x_\tau/\sqrt{\bar\alpha_\tau}$ the optimal denoiser is
\begin{equation}
\begin{aligned}
\mathbb{E}[x \mid x_\tau, a] &= \sum_{k} r_k(z_\tau)\, \big[(1-\lambda_\tau)\, z_\tau + \lambda_\tau\, \mu_k\big],\\
\lambda_\tau = \frac{s_\tau}{\sigma^2 + s_\tau}&, \quad r_k(z) \propto \pi_k\, \mathcal{N}\big(z;\, \mu_k, (\sigma^2 + s_\tau) I\big) ,
\end{aligned}
\label{eq:posterior}
\end{equation}
so the actions enter only through $\lambda_\tau \mu_k$ and the responsibilities $r_k$. At high noise $r_k \to \pi_k$ and the output is the conditional mean (Proposition~\ref{prop:limit}); once $s_\tau$ falls below the separation of the futures, $r_k$ selects the future nearest the noised goal, which is where the energy of $a$ drops in Fig.~\ref{fig:field}.

\begin{proposition}[Conditional channel]
\label{prop:channel}
Under Eq.~\ref{eq:posterior}, the gradient of the $x_0$ objective with respect to the conditional channel $\mu_k$ is scaled by $\lambda_\tau^2$, and that of the noise-prediction objective by $\lambda_\tau^2 \bar\alpha_\tau / (1-\bar\alpha_\tau)$. Both vanish as $\bar\alpha_\tau \to 1$: at low noise the denoiser can read the target from $x_\tau$ and receives no training signal for the actions. The noise-prediction weighting additionally multiplies the high-noise levels, where $\lambda_\tau \to 1$, by $\bar\alpha_\tau/(1-\bar\alpha_\tau) \to 0$.
\end{proposition}

The proposition prescribes how a joint-embedding predictor should be trained if its objective is to be a cost: the conditional channel is learned at high and intermediate noise, so training should concentrate there, and the $\epsilon$ parametrisation discounts exactly those levels, so the predictor should predict the clean target (Sec.~\ref{sec:analysis}).

\subsection{Horizon and the range of the cost}
\label{sec:horizon}

The predictor describes $p(x_{t+h} \mid c, a)$ for $h \le H$. A goal more than $H$ steps ahead is outside the support of every conditional the model has learned; there the energy varies with the actions only through how predictable the future is, and ranks the ground-truth chunk no better than chance. The range of the cost is therefore the training horizon (Sec.~\ref{sec:analysis}), extended by training longer with the same planner.

\subsection{Composition}
\label{sec:composition}

Since $E$ is a negative log-likelihood up to constants, conditionally independent goals multiply and energies add,
\begin{equation}
E(a \mid g_1, \dots, g_M) = \sum_m E(a \mid g_m),
\label{eq:product}
\end{equation}
the product of experts~\cite{hinton2002poe} in the latent space of the world model. The goals may be several views of one goal, the end states of several demonstrations, or samples from a conditional distribution over goals such as the language channel of Sec.~\ref{sec:model}. The chunk's own prior multiplies in as well, so planning is maximum a posteriori inference under $p(a \mid c, g_{1:M}) \propto p(a) \prod_m p_\theta(g_m \mid c, a)$,
\begin{equation}
a^\star = \arg\min_a \Big[ \sum_{m} E(a \mid g_m) - \log p(a) \Big],
\label{eq:map}
\end{equation}
with the planner's bounded Gaussian as $p(a)$. Scores of noised distributions do not add exactly at finite noise~\cite{du2023reduce}; we add energies rather than scores, under one noise draw shared by all experts and candidates, so the composition ranks candidates consistently where it is not an exact density.

\section{S-JEPA}
\label{sec:model}

S-JEPA is the joint-embedding world model that Sec.~\ref{sec:theory} prescribes: a score-based predictor whose objective is its planning cost, with four changes to the usual recipe, namely the parametrisation and noise schedule of Proposition~\ref{prop:channel}, a language channel in the action slot, joint training of the encoder, and a planner that evaluates the energy instead of rolling out (Fig.~\ref{fig:pipeline}).

All four share one network. The predictor is one transformer over three token groups: the $k$ context frames, whose tokens carry their frame's proprioception and action by channel concatenation as in DINO-WM~\cite{dinowm}; $h$ action tokens for the chunk; and the $P+1$ noised target tokens. A block mask lets context attend to earlier context, actions to context and actions, and targets to everything. A sinusoidal embedding of the noise level and a learned one of the horizon are added to every token, so one network serves all $h \le H$ and all levels.

Following Proposition~\ref{prop:channel}, the network predicts the clean target rather than the noise. Half of each batch draws its noise level from the upper third of the schedule, $\tau \in [0.7T, T)$, where the noised target does not reveal itself and the actions have to be read; the other half is uniform over all levels, which keeps the lower levels trained for sampling.

For language goals, a text embedding $u$ of the instruction, from the frozen CLIP ViT-B/32 text tower~\cite{radford2021clip}, takes the place of the chunk. It enters the action slot as a single token; the target tokens then see the context and the instruction, and the same network also models $p(x_{\mathrm{goal}} \mid c, u)$ (training details in Appendix~\ref{app:calvin}). At planning time $K$ goal latents are sampled from this conditional by DDIM in ten steps, once per instruction, and enter Eq.~\ref{eq:product} as goals,
\begin{equation}
E(a \mid u) = \sum_{k=1}^{K} E\big(a \mid g^{(k)}\big), \quad g^{(k)} \sim p_\theta(\cdot \mid c, u),
\label{eq:lang}
\end{equation}
so the instruction never has to be rendered as an image.

When the encoder is trained jointly, the denoising objective acquires a trivial minimiser: if all targets collapse to one point the noise can be read off the noised input and the loss vanishes. We train the encoder against an EMA copy of itself that produces the targets without gradient, as in JEPA~\cite{ijepa,bardes2024vjepa}, with a variance hinge on the online context tokens~\cite{bardes2022vicreg} (Appendix~\ref{app:joint}).

The planner solves Eq.~\ref{eq:map} for a chunk of $H$ actions by the cross-entropy method: $N$ candidates per round from a bounded diagonal Gaussian, $M$ elites, $R$ rounds. The energy uses a few noise levels in the content band and $J$ noise draws shared across candidates; the first $e$ actions of the mean are executed and the planner re-runs (Appendix~\ref{app:calvin}). Cameras are views of one state: each camera has its own predictor of this design, the goal is given in every camera, and the energies add by Eq.~\ref{eq:product}.

\section{Experiments}
\label{sec:exp}

The energy plans CALVIN from a goal image and from the instruction alone better than the latent distance of a predictor with the same architecture, data and budget, and plans the released NWM and the DINO-WM benchmarks without a rollout (Secs.~\ref{sec:calvin}--\ref{sec:maze}). Where the actions leave the future undetermined it ranks them by the probability of reaching the goal (Sec.~\ref{sec:stoch}). It does so because it is a likelihood of the goal whose resolution the noise level sets and whose range the training horizon sets, carried by a predictor trained where the actions are read (Sec.~\ref{sec:analysis}). The same planner, unchanged, then runs on a real robot arm (Sec.~\ref{sec:robot}).

\subsection{Setup}
\label{sec:setup}

CALVIN~\cite{calvin} provides 24 hours of teleoperated play with a Franka arm in four table-top scenes, 1\% of which carries a language annotation, and a fixed evaluation set of five-instruction chains in scene D; we use the $D\!\to\!D$ and $ABCD\!\to\!D$ splits. \emph{Windows}: the simulator is reset to the start of a language-annotated validation segment and the planner has 360 simulator steps to complete the task, judged by the CALVIN oracle; the goal is the frame at the end of the segment in each camera with its proprioception, or the instruction alone; all 1,011 validation segments. \emph{Chains}: the official 1,000 five-instruction sequences, each instruction given after the previous one succeeds. On the DINO-WM benchmarks~\cite{dinowm}, PointMaze, PushT and Wall, the protocol is the official one: 50 start--goal pairs with the goal five steps ahead, three rounds of model-predictive control, each a cross-entropy search over 300 candidates.

On CALVIN, frames of the static and the gripper camera are encoded once by the frozen V-JEPA~2 ViT-L/16 encoder~\cite{vjepa2} at 10~Hz, with one predictor per camera; the state is the end-effector pose and gripper width, and the action between two frames is the change of pose and the gripper command. Every predictor is trained on the same cached tokens with the same budget (six transformer layers of width 1024, 100k steps of batch 32), to a horizon of $0.5$ or $2$~s (Appendix~\ref{app:calvin}). The baseline follows the V-JEPA~2-AC recipe with the same architecture and inputs, trained to regress the target tokens and planned with the latent distance; its $L_2$ form, the better of the two norms on this data, is the one reported. A cross ablation plans the score-based predictor with the distance at $\tau \to T$, which Proposition~\ref{prop:limit} identifies with the baseline cost. On the DINO-WM benchmarks the encoder is the frozen DINOv2 ViT-S/14 of the released models, and the released DINO-WM predictors with their rollout-and-distance planner are the baselines. On navigation we read Eq.~\ref{eq:energy} from the released NWM CDiT-B~\cite{nwm} without training and compare with its official planner on its 100 evaluation trajectories (Appendices~\ref{app:dinowm},~\ref{app:nwm}).

The planner of Sec.~\ref{sec:model} uses $N = 96$ candidates, $R = 3$ rounds, $M = 16$ elites and two executed steps on CALVIN, with the energy at $\tau \in \{500, 800\}$, one shared noise draw per level and $K = 1$ sampled goal for language.

\subsection{CALVIN from images and from language}
\label{sec:calvin}

\begin{table}[t]
\centering
\footnotesize
\setlength{\tabcolsep}{2.6pt}
\begin{tabular}{llcccc}
\toprule
predictor, cost & goal & views & $D$ & $ABCD$ & s/plan \\
\midrule
regression, distance & image & 2 & 0.27 & 0.29 & 1.5 \\
S-JEPA, distance ($\tau\!\to\!T$) & image & 2 & 0.32 & -- & 1.5 \\
S-JEPA, energy & image & 1 & 0.15 & -- & 1.6 \\
S-JEPA, energy & image & 2 & 0.46 & 0.49 & 3.1 \\
S-JEPA, energy & text & 2 & 0.52 & \textbf{0.56} & 3.1 \\
regression $2$~s, distance & image & 2 & 0.31 & -- & 1.6 \\
S-JEPA $2$~s, energy & image & 2 & 0.57 & -- & 3.2 \\
S-JEPA $2$~s, energy & text & 2 & \textbf{0.63} & -- & 3.2 \\
\midrule
S-JEPA, energy, chains: length & text & 2 & 1.43 & 1.58 & -- \\
\quad first instruction & & & 0.60 & 0.64 & \\
\bottomrule
\end{tabular}
\caption{CALVIN with a frozen V-JEPA~2 encoder. Success over the 1,011 validation windows of scene $D$ (566 require a grasp) after training on $D$ or on $ABCD$, and planning time per call on one RTX~3090 running nothing else; views: static camera alone (1) or static and gripper cameras as a product of experts (2); text goals use $K = 1$ sampled goal; chains report the average length completed over the official 1,000 five-instruction sequences and, below, the success of the first instruction. Regression is the V-JEPA~2-AC recipe on the same tokens and budget. Higher is better.}
\label{tab:calvin}
\end{table}

Table~\ref{tab:calvin} is the main comparison: the same frozen encoder, cached data, training budget and planner, with only the training objective and its cost changed. With the energy of Eq.~\ref{eq:energy} the score-based predictor completes 46\% of the 1,011 windows from a goal image, against 27\% for the regression predictor with the latent distance and 32\% for the score-based predictor read at $\tau \to T$, its own latent-distance limit. Of the nineteen points, five come from the predictor and fourteen from the levels below the limit, and the fourteen fall where the predictor's sampled futures disagree: 22 points on the third of the windows where they differ most and 4 on the third where they differ least. Paired 95\% intervals are at most $\pm 3.4$ points (Appendix~\ref{app:calvin}). Of the windows, 566 need a grasp and 445 do not. The energy completes 69\% of the latter against 42\% for the latent distance, and 28\% of the grasp windows against 15\%. With the static camera alone it completes 15\% of all windows and 3\% of the grasp windows: the gripper camera enters as a second expert of Eq.~\ref{eq:product} and supplies the resolution that a grasp needs. Training on four times the data ($ABCD\!\to\!D$) adds three points and training to a longer horizon adds eleven: a predictor trained to $2$~s and read once completes 57\% at $3.2$~s per plan, against 31\% for the regression predictor trained to $2$~s, and ten of the eleven points come from the 551 windows whose goal pose lies more than 10~cm from the start, which the $0.5$-s predictor reaches only by replanning.

Given the instruction alone, with one goal sampled from the language channel, the same predictor completes 52\% of the windows, and the $2$-s predictor 63\%: the instruction is a better goal than the picture of the goal, because the sampled latent is a state the model finds reachable from the context, where the picture fixes details the planner cannot act on within the horizon. Several sampled goals compose as alternatives, not as a product (Appendix~\ref{app:calvin}).

\subsection{Goal views and the released NWM}
\label{sec:nav}

With the released Navigation World Model (CDiT-B)~\cite{nwm} and its 100 evaluation trajectories, the official planner samples 50-step trajectories, rolls the diffusion model, and ranks by the distance of the simulated endpoint to the goal. Reading the released model's denoising objective at the goal embedding, with no rollout and no further training, gives Eq.~\ref{eq:energy}. With one goal view the energy reaches an absolute trajectory error of 1.12 in 4.5~s per trajectory, against 1.107 in 188~s for the official planner (Tab.~\ref{tab:nwm}). With two goal views multiplied as experts it reaches 0.93 in 8.9~s, against 0.922 in 189~s, and with four views 0.85 in 18~s. At each number of views the energy reaches the official planner's error in a twentieth to a fortieth of its time.

\begin{table}[t]
\centering
\footnotesize
\setlength{\tabcolsep}{3.5pt}
\begin{tabular}{lcccc}
\toprule
NWM planner & goal views & ATE & RPE & s/traj. \\
\midrule
no planning (forward) & -- & 1.856 & 0.519 & -- \\
official, rollout + distance & 1 & 1.107 & 0.345 & 188 \\
official, rollout + distance & 2 & 0.922 & 0.326 & 189 \\
energy (Eq.~\ref{eq:energy}) & 1 & 1.12 & 0.331 & 4.5 \\
energy, product of goals & 2 & 0.93 & 0.309 & 8.9 \\
energy, product of goals & 4 & 0.85 & 0.301 & 18 \\
\bottomrule
\end{tabular}
\caption{Navigation planning with the released NWM, no training. Absolute and relative trajectory error (waypoint units) over the 100 trajectories, and planning time per trajectory on one RTX~3090 running nothing else, over the first five. Lower is better.}
\label{tab:nwm}
\end{table}

\subsection{The DINO-WM benchmarks}
\label{sec:maze}

On PointMaze the released DINO-WM predictor with its rollout-and-distance planner reaches the goal in 50 of 50 instances in 145 minutes on one RTX~3090. Our predictor on the same DINOv2 tokens, planned with the energy, reaches 48 of 50 in 5 minutes at the limit level $\tau = 980$ and 50 of 50 in 9 minutes at the two levels of CALVIN (Tab.~\ref{tab:maze}): the per-candidate rollout is gone. PushT and Wall follow: 0.90 against 0.92 in 50 minutes against 7.7 hours, and 0.90 against 0.92 in 15 minutes against 61. A regression predictor with the same time skip, read once with the latent distance, is as fast and reaches 0.61 and 0.70 (Appendix~\ref{app:dinowm}): the single pass gives the speed and the energy the success.

\begin{table}[t]
\centering
\footnotesize
\setlength{\tabcolsep}{4pt}
\begin{tabular}{llcc}
\toprule
environment & planner & success & time \\
\midrule
PointMaze & DINO-WM, rollout + distance & 1.00 & 145 min \\
 & S-JEPA, energy, $\tau = 980$ & 0.96 & 5 min \\
 & S-JEPA, energy, $J = 1$ & 1.00 & 9 min \\
PushT & DINO-WM, rollout + distance & 0.92 & 7.7 h \\
 & S-JEPA, energy, $J = 4$ & 0.90 & 50 min \\
Wall & DINO-WM, rollout + distance & 0.92 & 61 min \\
 & S-JEPA, energy, $J = 1$ & 0.90 & 15 min \\
\midrule
Stochastic & DINO-WM, rollout + distance & 0.38 & 58 min \\
Wall & S-JEPA, distance ($\tau \to T$) & 0.42 & 8 min \\
 & S-JEPA, energy, $J = 1$ & \textbf{0.86} & 15 min \\
\bottomrule
\end{tabular}
\caption{The DINO-WM benchmarks under the official protocol, and Wall with two door slots of which one, unseen, is open (Sec.~\ref{sec:stoch}). Same encoder, same 50 instances, one RTX~3090 each; the energy reads $\tau \in \{500, 800\}$ with $J$ noise draws unless a level is given. Higher success and lower time are better.}
\label{tab:maze}
\end{table}

The encoder can also be trained jointly with the predictor: with the variance hinge of Sec.~\ref{sec:model} it plans as well as the frozen one, 0.92 against 0.90 on PushT, where the encoder trained without the hinge collapses to 0.14 (Appendix~\ref{app:joint}).

\subsection{Futures the actions do not determine}
\label{sec:stoch}

Proposition~\ref{prop:limit} says what the distance loses: it scores the mean of the futures that a chunk allows, where the energy below the limit scores the goal under each of them. We test this on a variant of the DINO-WM Wall environment whose wall has two door slots, drawn alike, of which one is open in each episode. A chunk aimed through a slot ends beyond the wall or against it, and the goal lies just beyond the wall, between the slots (Appendix~\ref{app:dinowm}). Over 300 random chunks per instance, each executed under both door states, the energy separates the chunks that can reach the goal from those that cannot with an AUROC of 0.96, against 0.71 for the same predictor read at $\tau \to T$ and 0.68 for the rollout cost of a DINO-WM predictor trained on the same data. The distance prefers the chunk that stops against the wall opposite the goal: its single future is nearer to the goal than the mean of the two futures through a slot. The planners inherit the ordering, and the energy reaches the goal in 0.86 of 50 instances against 0.42 and 0.38 (Tab.~\ref{tab:maze}). The cost gap between the two chunks changes sign at $\tau \approx 640$, the crossing that Fig.~\ref{fig:field} draws in closed form. The lead follows the uncertainty: as the probability of the upper slot being the open one goes from $1$ to $0.9$, $0.7$ and $0.5$, the energy leads the distance by 0, 12, 31 and 44 points of success (Appendix~\ref{app:dinowm}).

\subsection{Representation analysis}
\label{sec:analysis}
Three properties of the energy underlie these results; we measure them on the static-camera predictor alone (Appendix~\ref{app:diag}): where in the noise schedule the actions are read, how far the cost sees, and how it tracks distance to the goal.

\begin{figure}[t]
\centering
\includegraphics[width=0.84\linewidth]{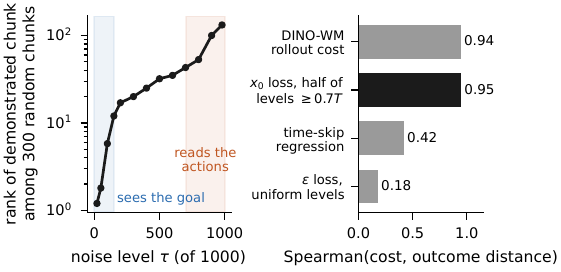}
\caption{The parametrisation decides whether the predictor learns a conditional channel. Left: rank of the demonstrated chunk among 300 random chunks when the energy of an $\epsilon$-parametrised predictor is read at one noise level (PushT, 12 instances; lower is better): the low levels identify the chunk because the noised goal reveals itself. Right: Spearman correlation between the cost of a random chunk and the true distance of its outcome to the goal (PushT, over 16 instances).}
\label{fig:per_t}
\end{figure}
\begin{figure}[t]
\centering
\includegraphics[width=0.82\linewidth]{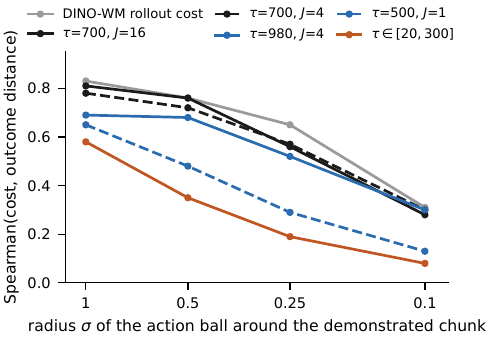}
\caption{The noise level is the resolution of the cost. Spearman correlation between cost and the true distance of the outcome for 100 chunks drawn in a ball of radius $\sigma$ around the demonstrated chunk (PushT, $x_0$ predictor at 25k steps). Ranking quality near the optimum rises with the noise level and with the number of noise draws $J$; the lowest levels order nothing. Higher is better.}
\label{fig:resolution}
\end{figure}
On PushT we rank 300 random action chunks by the energy read at a single noise level and record where the demonstrated chunk falls (Fig.~\ref{fig:per_t}). For a predictor trained with the noise-prediction loss the demonstrated chunk ranks 1.2 of 300 at $\tau = 20$ and 132 at $\tau = 980$: the low levels identify it because they see the goal, not because they read the actions. The direct test is the correlation between the energy and the true distance of the chunk's outcome to the goal, measured in the simulator. Under the noise-prediction loss with uniform levels this Spearman correlation is 0.06--0.25 depending on the levels. The same architecture with the $x_0$ loss and half of its levels above $0.7T$ reaches 0.95 after a quarter of the training, the value of the rollout planner's own cost (0.94); a regression predictor of the same architecture, read once with the latent distance, reaches 0.42. Inside a ball of radius $\sigma$ around the demonstrated chunk the ranking quality rises with the noise level (Fig.~\ref{fig:resolution}): at $\sigma = 0.5$ the correlation is 0.35 at the lowest levels, 0.59 at $\tau = 700$ and 0.76 with sixteen noise draws, against 0.76 for the rollout cost. The actions are read in the upper half of the schedule, as Proposition~\ref{prop:channel} says, and the expectation over noise draws, which the objective already contains, sets the resolution near the optimum.

\begin{figure}[t]
\centering
\includegraphics[width=0.84\linewidth]{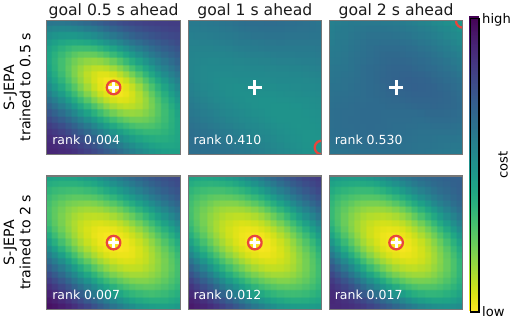}
\caption{The range of the cost follows the training horizon. Energy over a $21\times21$ grid of first actions ($\pm 6$~cm) around the demonstrated one (white cross; red circle: minimum), averaged over 50 CALVIN segments, for goals $0.5$, $1$ and $2$~s ahead. The predictor trained to $0.5$~s loses the goal beyond its horizon; the predictor trained to $2$~s keeps it. Rank: fraction of grid points cheaper than the demonstrated action (0 best, 0.5 chance).}
\label{fig:horizon}
\end{figure}
To measure the range of the cost (Sec.~\ref{sec:horizon}) we place the goal $0.5$, $1$ and $2$~s ahead of a demonstrated CALVIN state and perturb the first demonstrated action on a $21\times21$ grid of $\pm 6$~cm. The rank of the demonstrated action among the grid points tells whether the energy still points at the goal (Fig.~\ref{fig:horizon}). With the predictor trained to $0.5$~s the rank is 0.004 at $0.5$~s, 0.41 at $1$~s and 0.53 at $2$~s, and the latent-distance baseline trained to the same horizon follows the same curve (0.003, 0.39, 0.47). The predictor trained to $2$~s keeps the rank at 0.007, 0.012 and 0.017 across the three distances, against 0.012, 0.022 and 0.031 for the latent-distance baseline trained to $2$~s. The cost knows about the goal exactly as far as the predictor was trained to look, nothing in the planner had to change to extend it, and of the two $2$-s predictors the energy is sharper at every distance.

Along 100 planned image-goal CALVIN episodes of the static-camera predictor we log, at every step, the energy of the ``do nothing'' chunk and the true distance of the end effector to the goal pose. Over 5,561 steps the energy averages 0.25 within 3~cm of the goal against 0.51 beyond 25~cm (Appendix~\ref{app:diag}): a monotone reading of distance to the goal, which lets it stop the real robot.

\subsection{Real robot}
\label{sec:robot}
The planner of Sec.~\ref{sec:calvin} runs unchanged on a 6-DoF Seeed reBot Arm B601-RS. The predictor has the architecture and budget of Sec.~\ref{sec:setup}, trained on 29.5k DROID~\cite{droid} episodes with the frozen V-JEPA~2 encoder and the language channel, then fine-tuned once on 200 trajectories recorded on the arm; regression on the same data is the baseline. Over five tasks and 20 trials each, the energy completes 85 of 100 from a goal image against 60 for the latent distance, and 94 from the typed instruction alone (Tab.~\ref{tab:robot}; Appendix~\ref{app:robot}). One planning call takes 1.2~s on an RTX~4090.

\begin{figure}[t]
\centering
\includegraphics[width=\robotw\linewidth]{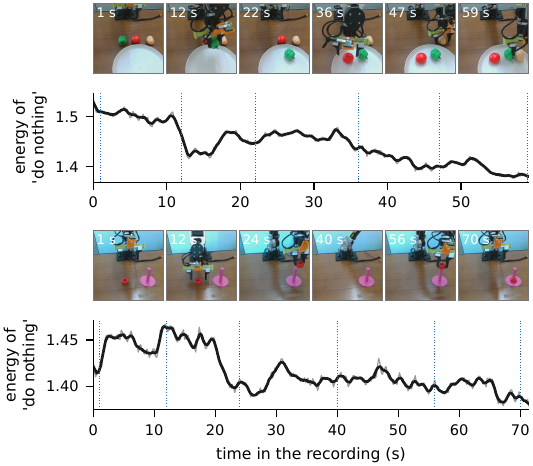}
\caption{The energy reads progress on the real arm. The arm places three objects on a plate (top) and a ring on a peg at its second attempt (bottom); frames at the dotted times. The energy of the ``do nothing'' chunk against the final frame, read at 5~Hz by the CALVIN-trained predictor, falls as the task advances (rank correlation with time $-0.88$ and $-0.74$).}
\label{fig:robot}
\end{figure}

\looseness=-1 An operator then types instructions one after another without a goal image: the arm stops when the energy of the ``do nothing'' chunk flattens, then the next is typed. The arm completes 11 of 15 four-instruction chains end to end, 53 of the 60 instructions, and its stop agrees with a human judge on 87 of 100 image and 92 of 100 text trials. The reading does not need in-domain training: the CALVIN predictor, which has seen only simulation, reads an energy on the real arm that falls as each object is placed (Fig.~\ref{fig:robot}).

\begin{table}[t]
\centering
\footnotesize
\setlength{\tabcolsep}{4pt}
\begin{tabular}{lccc}
\toprule
reBot arm, 20 trials per task & regression & S-JEPA & S-JEPA \\
 & image & image & text \\
\midrule
reach & 14 & \textbf{19} & \textbf{19} \\
push & 12 & 16 & \textbf{18} \\
drawer & 11 & 16 & \textbf{19} \\
button & 9 & 15 & \textbf{18} \\
pick-up & 14 & 19 & \textbf{20} \\
\midrule
4-instruction chains completed (15) & -- & -- & 11 \\
energy stop agrees with judge (100) & -- & 87 & 92 \\
\bottomrule
\end{tabular}
\caption{Real robot. Successes in 20 trials per task from a goal image (latent distance or energy) and from the typed instruction; four-instruction chains completed, of 15; energy stops that agree with the judge, of 100. Higher is better.}
\label{tab:robot}
\end{table}

\section{Conclusion}
\label{sec:conclusion}

\looseness=-1 A joint-embedding world model trained by denoising score matching contains its planning cost. The objective is a likelihood bound of the goal under the actions, and the latent distance that planners write by hand is its high-noise term. The noise level sets the resolution, the training horizon sets its range, and goals compose by multiplication. These statements fix how the predictor is trained and let one network plan from an image, several views, or language, one forward pass per candidate. The energy completes 63\% of CALVIN windows from the instruction against 31\% for latent distance and 94 of 100 real-robot trials against 60, reaches the NWM planner's error in a fortieth of its time, and matches the DINO-WM planner on PointMaze in a sixteenth. The representation bounds the cost resolution (Appendix~\ref{app:limitations}).

{
    \small
    \bibliographystyle{ieeenat_fullname}
    \bibliography{main}
}

\clearpage
\maketitlesupplementary
\setcounter{section}{0}\setcounter{table}{0}\setcounter{figure}{0}\setcounter{equation}{0}
\renewcommand{\thesection}{\Alph{section}}
\renewcommand{\thetable}{S\arabic{table}}
\renewcommand{\thefigure}{S\arabic{figure}}
\renewcommand{\theequation}{S\arabic{equation}}
\makeatletter
\renewcommand{\theHsection}{supp.\Alph{section}}
\renewcommand{\theHtable}{supp.\arabic{table}}
\renewcommand{\theHfigure}{supp.\arabic{figure}}
\renewcommand{\theHequation}{supp.\arabic{equation}}
\makeatother
\noindent This supplement gives the derivations behind the three propositions (Appendix~\ref{app:proofs}), the CALVIN data, models and protocols (Appendix~\ref{app:calvin}), the full diagnostic tables (Appendix~\ref{app:diag}), the DINO-WM benchmarks with the stochastic Wall environment and the NWM details (Appendices~\ref{app:dinowm} and \ref{app:nwm}), the joint training of the encoder (Appendix~\ref{app:joint}), the real-robot study (Appendix~\ref{app:robot}) and the limitations (Appendix~\ref{app:limitations}). Equation and table numbers of the form Eq.~1 or Tab.~1 refer to the main paper.
\section{Proofs and derivations}
\label{app:proofs}

\subsection{Notation}
\label{app:notation}
Targets $x \in \mathbb{R}^{(P+1)\times D}$ are the standardised tokens of a frame together with its proprioceptive token. The forward process is variance preserving: with $\beta_\tau$ linear in $\tau$ and $\bar\alpha_\tau = \prod_{s \le \tau}(1-\beta_s)$, a noised target is $x_\tau = \sqrt{\bar\alpha_\tau}\,x + \sqrt{1-\bar\alpha_\tau}\,\epsilon$ with $\epsilon \sim \mathcal{N}(0, I)$. We write $y = (c, a)$ for the condition (context frames with their actions, and the action chunk) and $p(x \mid y)$ for the distribution of the future representation given the condition. The signal-to-noise ratio of level $\tau$ is $\mathrm{SNR}_\tau = \bar\alpha_\tau / (1-\bar\alpha_\tau)$ and its logarithm $\ell_\tau=\log\mathrm{SNR}_\tau$. We write $\omega_\tau$ for the posterior-mean coefficient denoted $\lambda_\tau$ in Eq.~\ref{eq:posterior} of the main paper, distinguishing it from the energy weights $\lambda_\tau$ in Eq.~\ref{eq:energy}.

The three parametrisations of a denoiser are related by the identities of Ho et al.~\cite{ho2020ddpm}: an estimate $\hat{x}_0(x_\tau, \tau \mid y)$ of the clean target gives the noise estimate $\hat\epsilon = (x_\tau - \sqrt{\bar\alpha_\tau}\,\hat{x}_0)/\sqrt{1-\bar\alpha_\tau}$ and, by Tweedie's formula, the score estimate $\hat{s}(x_\tau \mid y) = -\hat\epsilon / \sqrt{1-\bar\alpha_\tau} = (\sqrt{\bar\alpha_\tau}\,\hat{x}_0 - x_\tau)/(1-\bar\alpha_\tau)$. The minimiser of the $x_0$ objective at level $\tau$ is the posterior mean $\mathbb{E}[x \mid x_\tau, y]$~\cite{vincent2011connection}, so the three losses share a minimiser and differ only in how levels are weighted:
\begin{equation}
\begin{aligned}
&\lVert \hat\epsilon - \epsilon \rVert^2 = \mathrm{SNR}_\tau \, \lVert \hat{x}_0 - x \rVert^2,\\
&\lVert \hat{s} - \nabla_{x_\tau}\log p_\tau(x_\tau \mid x) \rVert^2 = \frac{\lVert \hat\epsilon - \epsilon \rVert^2}{1-\bar\alpha_\tau}.
\end{aligned}
\label{eq:app:weights}
\end{equation}

\subsection{Proposition 1: the energy is a likelihood bound}
\label{app:prop1}
Song et al.~\cite{song2021maximum} (Theorem~3) show for the continuous-time version of the forward process with diffusion coefficient $g(t)$ that, for any data point $x$ and any score model $s_\theta$,
\begin{equation}
-\log p_\theta(x) \;\le\; \mathcal{J}_{\mathrm{DSM}}(\theta; g^2)(x) + C_{\mathrm{VP}}(x),
\label{eq:app:song}
\end{equation}
where $\mathcal{J}_{\mathrm{DSM}}(\theta; g^2)(x) = \tfrac12 \int_0^T g(t)^2\, \mathbb{E}_{x_t \mid x}\lVert s_\theta(x_t, t) - \nabla_{x_t}\log p_t(x_t \mid x)\rVert^2\, dt$ is the denoising objective at $x$ with the likelihood weighting $g^2$, $p_\theta$ is the marginal of the reverse process started from the prior $\pi$, and $C_{\mathrm{VP}}(x)$ retains the terminal cross-entropy and the forward-process normalisation terms of their pointwise bound. Writing $q_t(\cdot\mid x)$ for the forward kernel, $b_t(u)$ for its drift and $g(t)$ for its diffusion coefficient, these terms are
\[
\begin{aligned}
C_{\mathrm{VP}}(x)={}&-\mathbb{E}_{q_T}\log\pi(x_T)\\
&-\frac12\int_0^T\mathbb{E}_{q_t}\big[g(t)^2
\|\nabla\log q_t(x_t\mid x)\|^2\\
&\hspace{32mm}+2\nabla\!\cdot b_t(x_t)\big]\,dt .
\end{aligned}
\]
The endpoint terms are evaluated together under the limiting or regularisation convention of the full bound. For the variance-preserving process $b_t(u)=-\beta(t)u/2$, its divergence is independent of the action. The proof conditions everything on $y$ without change: the forward process does not depend on $y$, the reverse process uses $s_\theta(x_t, t \mid y)$, and the inequality holds pointwise in $y$. The normalisation terms are the same for every chunk $a$ when the goal, forward schedule, latent coordinates and prior are fixed and independent of $a$. Let $E_{\mathrm{cont}}(a\mid g)$ denote the continuous likelihood-weighted denoising objective, with its boundary terms retained. Conditioning Eq.~\ref{eq:app:song} on $(c,a)$ gives
\begin{equation}
E_{\mathrm{cont}}(a \mid g) \;\ge\; -\log p_\theta(g \mid c, a) - C(g),
\label{eq:app:bound}
\end{equation}
with $C(g)$ independent of $a$, which is the first statement of Proposition~\ref{prop:bound}. In the $x_0$ parametrisation $s_\theta - \nabla \log p_t(x_t \mid x) = \sqrt{\bar\alpha_t}\,(f_\theta - x)/(1-\bar\alpha_t)$, and for the variance-preserving process $g(t)^2\, \bar\alpha_t/(1-\bar\alpha_t)^2\, dt = -d\gamma$ with $\gamma = \bar\alpha_t/(1-\bar\alpha_t)$, so $\mathcal{J}_{\mathrm{DSM}}(\theta; g^2)(x) = \tfrac12\int_0^\infty \mathbb{E}_\epsilon \lVert f_\theta(x_\gamma) - x\rVert^2\, d\gamma$, the form of Eq.~\ref{eq:bound}; a quadrature approximation over the levels uses the weights $\lambda^\star_\tau = \tfrac12(\gamma_{\tau-1} - \gamma_\tau)$. Eq.~\ref{eq:residual} follows by inserting $f_\theta = (g_\tau + (1-\bar\alpha_\tau)\, s_\theta)/\sqrt{\bar\alpha_\tau}$ and $g_\tau = \sqrt{\bar\alpha_\tau}\, g + \sqrt{1-\bar\alpha_\tau}\,\epsilon$ into $f_\theta - g$. For the mixture of Eq.~\ref{eq:posterior}, each component is the Gaussian case below and the posterior over components given $z_\tau = x_\tau/\sqrt{\bar\alpha_\tau} = x + \sqrt{s_\tau}\,\epsilon$ is $r_k(z_\tau) \propto \pi_k\, \mathcal{N}(z_\tau; \mu_k, (\sigma^2 + s_\tau) I)$. Minimising $E_{\mathrm{cont}}$ over $a$ minimises an upper bound on the negative conditional log-likelihood of the goal; tightness requires an exact compatible score and matching terminal prior, under the regularity and endpoint conditions of the cited result.

For the second statement, the monotone-weighting result of Kingma and Gao~\cite{kingma2023understanding} applies to the complete effective weight in log-SNR, including the sampling density and change-of-variables Jacobian, with the endpoint terms retained. Let $\ell=\ell_\tau$, $q(\tau)$ be the level-sampling density and $\rho(\ell)=q(\tau(\ell))|d\tau/d\ell|$. Up to a positive overall scale, an unweighted $x_0$ training loss induces $w(\ell)=\rho(\ell)e^{-\ell}$ in the noise-prediction integral. When $w(\ell)$ is non-increasing in $\ell$, the continuous objective corresponds to a negative ELBO averaged over forward-noised targets, with the endpoint terms retained.

\paragraph{Continuous objective and planning score.} The likelihood bound applies to the continuous objective with the specified weighting and boundary conditions. Its density is the conditional marginal of the reverse SDE, under the regularity conditions of the cited theorem; equality additionally requires an exact compatible score and matching terminal prior. The planner evaluates a finite-level, finite-sample approximation as a likelihood-motivated planning cost. The bound concerns $E_{\mathrm{cont}}$, while the finite score and its candidate ordering are assessed empirically. In CALVIN, the two-level, one-draw score has a Spearman rank correlation of 0.97 with a sixteen-level, four-draw quadrature (Appendix~\ref{app:pertask}).

\paragraph{Computational cost.} With $L$ noise levels, $J$ draws per level and $V$ independently evaluated views, each candidate requires $LJV$ single-level, single-view denoiser evaluations. Default CALVIN uses $L=2$, $J=1$ and two independent camera predictors ($V=2$), giving four equivalent evaluations per candidate. Batching groups these evaluations into hardware calls; the call count depends on the batching configuration.

\paragraph{Truncated bound.} The levels $\tau \ge \tau_{\min}$ of the forward process are a diffusion of the noised target $x_{\tau_{\min}}$, and the score model restricted to them defines a density $p_{\theta,\tau_{\min}}(\cdot \mid c, a)$ of that noised target. Applying Eq.~\ref{eq:app:song} to this diffusion, and using that denoising score matching against $x_{\tau_{\min}}$ and against $x$ differ by a term that depends on neither $\theta$ nor $a$~\cite{vincent2011connection},
\begin{equation}
\begin{aligned}
&\mathbb{E}_{\epsilon}\big[{-}\log p_{\theta,\tau_{\min}}(g_{\tau_{\min}} \mid c, a)\big] \\
&\qquad \le\; \frac12 \int_0^{\gamma_{\tau_{\min}}}\! E_{\tau(\gamma)}(a \mid g)\, d\gamma + C .
\end{aligned}
\label{eq:app:trunc}
\end{equation}
The energy over the levels above $\tau_{\min}$ therefore bounds the likelihood of the goal known to within $\sqrt{s_{\tau_{\min}}}$ per standardised dimension, the tolerance the planner sets by its lowest level. For planning, the integral is approximated using levels $\{500,800\}$ and one noise draw per level. This score is highly rank-correlated with the denser quadrature (Spearman $\rho=0.97$; Appendix~\ref{app:pertask}).

\paragraph{One Gaussian future.} For $x \mid c, a \sim \mathcal{N}(\mu(a), \sigma^2 I)$, Eq.~\ref{eq:posterior} gives $f(g_\tau) - g = \omega_\tau(\mu(a) - g) + (1-\omega_\tau)\sqrt{s_\tau}\,\epsilon$, so $E_\tau(a \mid g) = \omega_\tau^2 \lVert \mu(a) - g\rVert^2$ plus a term independent of $a$: every level orders chunks by the latent distance. The energy and the distance can differ only when the futures a chunk allows are not one Gaussian, and the difference grows with their separation (Appendix~\ref{app:stoch}).

\subsection{Proposition 2: the latent distance is the high-noise limit}
\label{app:prop2}
At level $\tau$ the optimal denoiser is the posterior mean $f^\star(x_\tau, \tau \mid y) = \mathbb{E}[x \mid x_\tau, y]$. As $\bar\alpha_\tau \to 0$ the likelihood $p(x_\tau \mid x) = \mathcal{N}(\sqrt{\bar\alpha_\tau}\,x, (1-\bar\alpha_\tau) I)$ becomes constant in $x$, the posterior converges to the prior $p(x \mid y)$, and $f^\star \to \mathbb{E}[x \mid y]$, a quantity that does not depend on the noise draw. The energy at that level is then
\begin{equation}
E_\tau(a \mid g) = \mathbb{E}_\epsilon \lVert f^\star - g \rVert^2 \;\to\; \lVert \mathbb{E}[x \mid c, a] - g \rVert^2 .
\label{eq:app:limit}
\end{equation}
A predictor trained to regress $x$ from $y$ with the squared loss also converges to $\mathbb{E}[x \mid y]$, so the latent-distance cost of a regression world model and the high-noise term of the energy are the same function of $(c, a, g)$. With the $L_1$ loss the regressor converges to the conditional median; the argument is unchanged for a Laplace rather than Gaussian limit. In the model we train, $\bar\alpha_{700} \approx 7\times10^{-3}$: at $\tau = 700$ the noised goal keeps a signal amplitude $\sqrt{\bar\alpha_{700}} \approx 0.08$ and the ranking still depends on the noise draws (Fig.~\ref{fig:resolution}), whereas at $\tau = 980$ the amplitude is $0.008$ and the energy is numerically this limit on every candidate.

\subsection{Proposition 3: which levels train the conditional channel}
\label{app:prop3}
Take the Gaussian model $x \mid y \sim \mathcal{N}(\mu_y, \sigma^2 I)$, where the condition acts only through the mean. The posterior of $x$ given $x_\tau$ is Gaussian with mean
\begin{equation}
\begin{aligned}
\mathbb{E}[x \mid x_\tau, y] &= (1-\omega_\tau)\,\frac{x_\tau}{\sqrt{\bar\alpha_\tau}} + \omega_\tau\,\mu_y,\\
\omega_\tau &= \frac{(1-\bar\alpha_\tau)/\bar\alpha_\tau}{\sigma^2 + (1-\bar\alpha_\tau)/\bar\alpha_\tau} ,
\end{aligned}
\label{eq:app:posterior}
\end{equation}
by completing the square in $\log p(x \mid y) + \log p(x_\tau \mid x)$. Now let the network implement this posterior mean with a learned estimate $\hat\mu_y$ of the conditional mean, which is the only place the condition enters. The $x_0$ objective at level $\tau$ is $\ell_{x_0} = \mathbb{E}\lVert (1-\omega_\tau)\, x_\tau/\sqrt{\bar\alpha_\tau} + \omega_\tau\,\hat\mu_y - x \rVert^2$, whose gradient with respect to $\hat\mu_y$ is $2\omega_\tau\, \mathbb{E}[(1-\omega_\tau)\, x_\tau/\sqrt{\bar\alpha_\tau} + \omega_\tau\hat\mu_y - x]$ and whose curvature in $\hat\mu_y$ is $2\omega_\tau^2$. The $\epsilon$ objective at the same level equals $\mathrm{SNR}_\tau\,\ell_{x_0}$ by Eq.~\ref{eq:app:weights}, so its curvature in $\hat\mu_y$ is $2\omega_\tau^2\,\bar\alpha_\tau/(1-\bar\alpha_\tau)$. These are the scalings stated in Proposition~\ref{prop:channel}. As $\bar\alpha_\tau \to 1$, $\omega_\tau \to 0$ and both objectives stop constraining $\hat\mu_y$: the network recovers $x$ from $x_\tau$ and never has to read $y$. As $\bar\alpha_\tau \to 0$, $\omega_\tau \to 1$; the $x_0$ objective constrains $\hat\mu_y$ fully, while the $\epsilon$ objective multiplies that constraint by $\mathrm{SNR}_\tau \to 0$. With uniform sampling of $\tau$ the $\epsilon$ objective therefore spends most of its gradient where the conditional channel is not needed, which is the mechanism behind the rank correlations of 0.06--0.25 ($\epsilon$, uniform) against 0.95 ($x_0$, half of the levels above $0.7T$) in Sec.~\ref{sec:analysis}. The argument uses a Gaussian conditional; for a multimodal $p(x \mid y)$ the posterior mean is a $\omega$-weighted combination of the modes and the same scaling applies to each.

\subsection{Horizon and the range of the cost}
\label{app:horizon}
The predictor is trained on pairs $(y, x_{t+h})$ with $h \le H$, so $p_\theta(x \mid c, a)$ is an estimate of the distribution of representations at most $H$ steps ahead. For a goal $g$ that the system can reach only in $h' > H$ steps, no chunk makes $g$ a typical sample of $p(x_{t+h} \mid c, a)$; the energy then compares, across chunks, how well the model predicts its own $H$-step future when that future is far from $g$, which varies with the predictability of the future under the chunk and not with the direction to the goal. This is the behaviour measured in Fig.~\ref{fig:horizon}: the demonstrated action ranks at chance among perturbations once the goal is beyond the horizon, and at the bottom of the landscape within it. Training to a longer $H$ enlarges the set of goals that are typical under some chunk, which is the only change needed to extend the range.

\subsection{Composition}
\label{app:composition}
If goals $g_1, \dots, g_M$ are conditionally independent given $(c, a)$, $\log p(g_{1:M} \mid c, a) = \sum_m \log p(g_m \mid c, a)$, and Eq.~\ref{eq:app:bound} applied to each term gives the sum of continuous objectives as a bound on the joint negative log-likelihood, up to a constant independent of $a$. Du et al.~\cite{du2023reduce} show that for the noised distributions $p_\tau$ the score of a product is not the sum of the scores of the factors at $\tau > 0$, so sampling from a composition requires correction. The planner ranks chunks by the sum of the factors' energies, each evaluated with the same noise draw $\epsilon$ for every chunk and every factor. In the high-noise limit of Proposition~\ref{prop:limit}, each energy is a squared distance and their sum is the squared distance to the precision-weighted mean of the goals, up to a candidate-independent term. At finite noise, the experiments evaluate the summed score using common noise draws across candidates. The action prior of Sec.~\ref{sec:composition} is implemented by sampling standardised actions from a Gaussian clipped at three standard deviations, which keeps every candidate inside the range the predictor was trained on.

\section{CALVIN: data, models, protocol}
\label{app:calvin}

\subsection{Cached representation}
\label{app:cache}
CALVIN~\cite{calvin} records play at 30~Hz. We keep every third frame of the static and of the gripper camera (10~Hz), resize it to 256~px and encode it once with the frozen V-JEPA~2 ViT-L/16 encoder~\cite{vjepa2}, which yields $16\times16 = 256$ tokens of 1024 channels per frame; the cache stores them in half precision. The proprioceptive state is the seven-dimensional end-effector pose (position, Euler orientation, gripper width) of \texttt{robot\_obs}; the action between two cached frames is the change of the first six coordinates and the gripper command $\pm1$ of the last recorded step, standardised per dimension over the training split. Token standardisation statistics (per channel over tokens, and per dimension over the proprioceptive token) are estimated once from 60 batches at the start of training and stored with the model. After caching, the $D$ split holds 170,703 training frames (4.7~h of play in 31 recorded episodes) and 33,008 validation frames (0.9~h); $ABCD$ holds 769,098 training frames (21.4~h) and shares the validation frames. Language annotations (\texttt{auto\_lang\_ann}) give 5,124 training and 1,011 validation windows in $D$ (22,966 training windows in $ABCD$), each a start and end frame with an instruction; instructions are embedded once by the frozen CLIP ViT-B/32 text tower~\cite{radford2021clip} into a table.

\subsection{Predictors}
\label{app:predictors}
There is one predictor per camera, trained on that camera's tokens with the shared proprioceptive state and actions. All predictors share the architecture of Sec.~\ref{sec:model}: six transformer layers of width 1024 with 16 heads, context $k \in \{1, 2, 3\}$ frames drawn per batch, horizon $h$ drawn uniformly in $\{1, \dots, H\}$ per batch, 257 target tokens (256 visual and one proprioceptive). Context tokens carry a 10-dimensional proprioception embedding and a 10-dimensional action embedding by concatenation; the chunk enters as $h$ action tokens through the same action encoder. Training: 100k steps of batch 32 (20 epochs of 5,000 steps), AdamW with $\beta = (0.9, 0.95)$, weight decay $0.01$, peak learning rate $5\times10^{-4}$ after 1,000 warm-up steps and cosine decay to 10\%, gradient norm clipped at 1.0, bfloat16 autocast. S-JEPA uses the $x_0$ parametrisation with $T = 1000$ levels, linear $\beta$ from $10^{-4}$ to $0.02$, and the mixed level sampler (half uniform, half uniform on $[700, 1000)$); the model with the language channel adds one language batch every two steps with its own optimiser step. The regression predictor uses the same inputs and outputs with the target standardised per channel and the squared loss; Table~\ref{tab:app:recipes} lists the other regression recipes, including the $L_1$ form that is the literal V-JEPA~2-AC loss. The $2$-s predictors are trained with $H = 20$ and are otherwise identical. The peak learning rate of every predictor is the one of $\{2\times10^{-4}, 5\times10^{-4}\}$ with the lower validation loss at 100k steps: $5\times10^{-4}$ for all predictors but the $2$-s regression predictor, which takes $2\times10^{-4}$ with 2,000 warm-up steps.

\begin{table}[h]
\centering
\footnotesize
\begin{tabular}{lc}
\toprule
regression recipe ($D\!\to\!D$, image goal, two views) & success \\
\midrule
$L_1$, per-channel standardisation & 0.05 \\
$L_1$, per-token layer norm & 0.06 \\
$L_2$, per-channel standardisation (reported) & 0.27 \\
$L_2$, per-token layer norm & 0.09 \\
\bottomrule
\end{tabular}
\caption{Regression recipes on the same tokens and budget, 1,011 windows. The $L_2$ per-channel form is the baseline of the main text.}
\label{tab:app:recipes}
\end{table}

\paragraph{Language channel.} The text embedding of the instruction is projected into the action slot as a single token with its own position embedding and a horizon id reserved for unknown horizons. Training interleaves one language batch every two dynamics batches, drawn from the language-annotated windows of play (context at the start of the window, target at its end, and the instruction).

\subsection{Windows protocol}
\label{app:windows}
A window is a language-annotated validation segment. The simulator is reset to the segment's recorded start state (robot and scene), and the planner controls the arm for at most 360 simulator steps, the per-instruction budget of the official chain benchmark; success is judged by the CALVIN task oracle for the window's task. The goal is either the segment's end frame in each camera with its proprioceptive state (image goal) or the instruction alone (language goal). Each 10-Hz action from the planner is executed as three relative actions of the simulator, each a third of the pose change with the gripper command repeated. Every row is run on all 1,011 validation windows. Replaying a window's recorded actions, with no model, succeeds in 0.82 of windows at 30~Hz and 0.68 at the planner's 10-Hz resolution.

\subsection{Chains protocol}
\label{app:chains}
The official benchmark~\cite{calvin} provides 1,000 sequences of five instructions in scene D. We run all of them: the arm starts from the recorded initial state, each instruction is given to the planner in turn, the oracle decides success, and the next instruction is given only after success, with 360 simulator steps per instruction. We report the average number of instructions completed and the success of the first instruction (Tab.~\ref{tab:calvin}); the planner completes one to five instructions in 0.60, 0.37, 0.22, 0.14 and 0.10 of the $D\!\to\!D$ chains. Language goals are sampled once per instruction with $K = 1$.

\subsection{Planner settings}
\label{app:planner}
Cross-entropy method with $N = 96$ candidates, $R = 3$ rounds, $M = 16$ elites, mean and standard deviation updated with momentum $0.5$, candidates clipped to $\pm3$ in standardised action units, warm start by shifting the previous mean by the two executed steps; horizon $H = 5$ at 10~Hz (or $H = 20$ with four piecewise-constant knots); energy at $\tau \in \{500, 800\}$, a two-point quadrature of the truncated bound of Eq.~\ref{eq:app:trunc} with $\tau_{\min} = 500$, and one noise draw per level, shared by all candidates and both cameras, computed in bfloat16 in chunks of 256 candidates. With two cameras the energies of the two predictors add; the latent-distance rows add the two distances. Planning time is wall-clock per planning call, measured on the first 20 windows of each row on an otherwise idle RTX~3090.

\subsection{Per-task breakdown and ablations}
\label{app:pertask}
The 1,011 windows are language-annotated segments of the 34 CALVIN tasks: 566 need a grasp or start with a block in the gripper (lift, rotate, stack, unstack, place), and 445 do not (pushing a block, 253; moving the slider, 71; opening or closing the drawer, 51; switching the LED or the light bulb, 70). Table~\ref{tab:app:pertask} splits the rows of Tab.~\ref{tab:calvin} along these families. The energy leads the latent distance in every family, and the gripper camera moves the grasp windows from 17 to 156. The image and text goals of the same predictor succeed on 391 common windows, 74 further windows from the image alone and 135 from the text alone. Rolling the $0.5$-s predictor's conditional mean four times and reading the energy at the end completes 40\% at $7.6$~s per plan, against 57\% at $3.2$~s for the predictor trained to $2$~s. Table~\ref{tab:app:ablate} varies the levels and the noise draws of the energy and the composition of sampled language goals. Four sampled goals with their energies multiplied complete 47\%, below one goal: at high noise the energies of several goals add to the distance to their mean (Proposition~\ref{prop:limit}), and the mean of several samples of one instruction is a blurred goal. Read as alternatives, with the soft minimum of the four energies as the cost, the same samples complete 55\%. Paired bootstrap intervals (10,000 resamples of the windows) of the differences quoted in Sec.~\ref{sec:calvin} are $19 \pm 3.4$ points (energy against regression), $14 \pm 3.1$ (energy against its high-noise limit), $6 \pm 2.9$ (text against image), $11 \pm 3.2$ ($2$~s against $0.5$~s) and $32 \pm 3.3$ (the $2$-s predictor from text against the $2$-s regression predictor from the image). Over the 96 candidates of the first planning call of each window, the two-level, one-draw energy and a quadrature of Eq.~\ref{eq:app:trunc} over 16 levels in $[500, 980]$ with four draws have a Spearman correlation of 0.97; the quadrature costs 32 times the forward passes.

\paragraph{Where the gains fall.} Table~\ref{tab:app:where} splits the windows two ways. The first is by the spread of the predictor's futures: at the start of each window we draw eight samples of the $0.5$-s predictor under the demonstrated chunk and take the variance across samples, averaged over tokens of both cameras; the windows are cut into thirds by this spread. The energy and its high-noise limit complete nearly the same windows where the samples agree and part where they differ, the behaviour of Appendix~\ref{app:prop1} for one and for several futures. The spread is largest in the grasp windows, where the static view leaves open whether the fingers hold the block. The second split is by the distance between the end-effector poses of the start and the goal frame. The $0.5$-s and the $2$-s predictor complete the near windows alike; the $2$-s predictor adds the far ones, which the $0.5$-s predictor completes only when replanning brings the goal within its range.

\begin{table}[h]
\centering
\footnotesize
\setlength{\tabcolsep}{3pt}
\begin{tabular}{lccc}
\toprule
spread of futures & low third & mid third & high third \\
 & (337) & (337) & (337) \\
\midrule
S-JEPA, distance ($\tau\!\to\!T$) & 150 & 108 & 66 \\
S-JEPA, energy & 163 & 162 & 140 \\
\midrule
start-to-goal distance & \multicolumn{2}{c}{$\le 10$~cm (460)} & $> 10$~cm (551) \\
\midrule
regression, distance & \multicolumn{2}{c}{205} & 68 \\
regression $2$~s, distance & \multicolumn{2}{c}{212} & 101 \\
S-JEPA, energy & \multicolumn{2}{c}{322} & 143 \\
S-JEPA $2$~s, energy & \multicolumn{2}{c}{331} & 245 \\
\bottomrule
\end{tabular}
\caption{Windows completed from a goal image ($D\!\to\!D$, two views), split by the spread of the predictor's sampled futures and by the distance from start to goal; the number of windows in each group is in parentheses. Higher is better.}
\label{tab:app:where}
\end{table}

\begin{table}[h]
\centering
\footnotesize
\setlength{\tabcolsep}{3pt}
\begin{tabular}{lc}
\toprule
energy read by the planner ($D\!\to\!D$, two views) & success \\
\midrule
image goal, $\tau = 980$ (the limit of Proposition~\ref{prop:limit}) & 0.32 \\
image goal, $\tau = 700$ & 0.39 \\
image goal, $\tau \in \{500, 800\}$ (reported) & 0.46 \\
image goal, $\tau \in \{300, 500, 700, 900\}$ & 0.47 \\
image goal, $\tau \in \{500, 800\}$, $J = 4$ draws per level & 0.47 \\
image goal, 16 levels in $[500, 980]$, $J = 4$ & 0.47 \\
\midrule
text goal, $K = 1$ (reported) & 0.52 \\
text goal, $K = 4$, product & 0.47 \\
text goal, $K = 4$, soft minimum & 0.55 \\
\bottomrule
\end{tabular}
\caption{Levels, noise draws and goal composition, 1,011 windows. Higher is better.}
\label{tab:app:ablate}
\end{table}

\begin{table}[h]
\centering
\footnotesize
\setlength{\tabcolsep}{1.6pt}
\begin{tabular}{llcccccc}
\toprule
predictor, cost & goal & push & slider & drawer & LED & grasp & all \\
 & & (253) & (71) & (51) & (70) & (566) & (1011) \\
\midrule
\multicolumn{8}{l}{$D\!\to\!D$} \\
regression, dist. & image & 110 & 22 & 18 & 38 & 85 & 273 \\
S-JEPA, dist. ($\tau\!\to\!T$) & image & 124 & 30 & 22 & 40 & 108 & 324 \\
S-JEPA, energy, 1 view & image & 68 & 26 & 11 & 30 & 17 & 152 \\
S-JEPA, energy & image & 172 & 52 & 36 & 49 & 156 & 465 \\
S-JEPA, energy & text & 185 & 50 & 38 & 55 & 198 & 526 \\
regression $2$~s, dist. & image & 118 & 30 & 22 & 37 & 106 & 313 \\
S-JEPA $2$~s, energy & image & 190 & 62 & 41 & 50 & 233 & 576 \\
S-JEPA $2$~s, energy & text & 200 & 60 & 42 & 56 & 279 & 637 \\
S-JEPA, rolled $4{\times}0.5$~s & image & 150 & 40 & 30 & 44 & 140 & 404 \\
\midrule
\multicolumn{8}{l}{$ABCD\!\to\!D$} \\
regression, dist. & image & 115 & 24 & 20 & 39 & 95 & 293 \\
S-JEPA, energy & image & 178 & 54 & 38 & 51 & 174 & 495 \\
S-JEPA, energy & text & 192 & 53 & 40 & 57 & 224 & 566 \\
\bottomrule
\end{tabular}
\caption{CALVIN windows completed by task family, for the rows of Tab.~\ref{tab:calvin}; the number of windows in each family is in parentheses. Push includes pushing a block into the drawer; grasp covers lift, rotate, stack, unstack and place, the windows that need a grasp or start with a block in the gripper. Two views unless noted.}
\label{tab:app:pertask}
\end{table}

\section{Representation diagnostics: protocols and full tables}
\label{app:diag}
Section~\ref{sec:analysis} measures three properties of the energy on the static-camera predictor alone; this appendix gives the protocol and the full table behind each, in the same order: where in the noise schedule the actions are read (Figs.~\ref{fig:per_t} and~\ref{fig:resolution}), how far the cost sees (Fig.~\ref{fig:horizon}), and how it follows the distance to the goal (Fig.~\ref{fig:e0}).

\subsection{Per-level information and parametrisation (PushT)}
\label{app:pert}
On PushT (DINO-WM data and encoder, Appendix~\ref{app:dinowm}) we take 12--16 validation instances, draw 300 random action chunks of five steps, add the demonstrated chunk, and rank all 301 by the energy at one level. For the $\epsilon$-parametrised predictor trained with uniform levels the demonstrated chunk's mean rank is 1.2 at $\tau{=}20$, 5.8 at 100, 20 at 300, 32 at 500, 43 at 700, 100 at 900 and 132 at 980 (lower is better). The outcome test executes each random chunk in the simulator and correlates the cost with the final distance to the goal state: $\epsilon$, uniform levels: 0.06--0.25 across level sets; $x_0$ with half of the levels above $0.7T$: 0.948 (levels 20--150), 0.938 (per-level standardised 50--900), 0.887 (300--700); the released DINO-WM rollout cost: 0.945; a regression predictor with the same time-skip architecture: 0.42.

\subsection{Energy landscapes (Fig.~\ref{fig:horizon})}
\label{app:landscape}
For 50 language-annotated validation segments we draw a start frame inside the segment with $k$ context frames before it and take the goal $\Delta \in \{5, 10, 20\}$ frames ($0.5$, $1$, $2$~s) ahead; the demonstrated chunk and the goal may run past the annotated end of the segment but stay inside the same episode. The predictor is queried with $h = \Delta$ when $\Delta \le H$ and with $h = H$ otherwise. The first demonstrated action's $(x, y)$ change is replaced by each point of a $21\times21$ grid spanning $\pm6$~cm, the remaining actions are kept, and the energy (or latent distance) of every grid point is computed with the same noise draw. We report the fraction of grid points with lower cost than the demonstrated action (rank), the distance from the demonstrated action to the minimum (cm), and the fraction of segments whose minimum lies within 2~cm (Table~\ref{tab:app:landscape}). Prediction itself reaches as far under either objective: the $2$-s S-JEPA predictor reduces the $L_1$ error of the target tokens from $0.70$ (copying the last frame) to $0.40$ at $2$~s, and its prediction moves by $0.48$ when the chunk is replaced by a random one, against $0.17$ at one step; the regression predictor trained to $2$~s at the same budget reaches $0.42$ and $0.45$ on the same two measures. The two objectives build predictors of the same reach and differ in the cost that can be read from them.

\begin{table}[h]
\centering
\footnotesize
\setlength{\tabcolsep}{4pt}
\begin{tabular}{llccc}
\toprule
predictor / cost & goal & rank & argmin (cm) & $\le 2$ cm \\
\midrule
regression $0.5$ s / distance & $0.5$ s & 0.003 & 0.4 & 1.00 \\
 & $1$ s & 0.385 & 6.1 & 0.08 \\
 & $2$ s & 0.467 & 6.6 & 0.02 \\
S-JEPA $0.5$ s / energy & $0.5$ s & 0.004 & 0.5 & 0.98 \\
 & $1$ s & 0.410 & 6.2 & 0.04 \\
 & $2$ s & 0.530 & 7.2 & 0.02 \\
regression $2$ s / distance & $0.5$ s & 0.012 & 0.8 & 0.98 \\
 & $1$ s & 0.022 & 0.9 & 0.94 \\
 & $2$ s & 0.031 & 1.2 & 0.84 \\
S-JEPA $2$ s / energy & $0.5$ s & 0.007 & 0.6 & 0.98 \\
 & $1$ s & 0.012 & 0.8 & 0.96 \\
 & $2$ s & 0.017 & 0.9 & 0.90 \\
\bottomrule
\end{tabular}
\caption{Energy landscapes on CALVIN validation play, 50 segments per row. Rank: fraction of the 441 grid points with lower cost than the demonstrated action (0 best, 0.5 chance).}
\label{tab:app:landscape}
\end{table}

\begin{figure}[h]
\centering
\includegraphics[width=\linewidth]{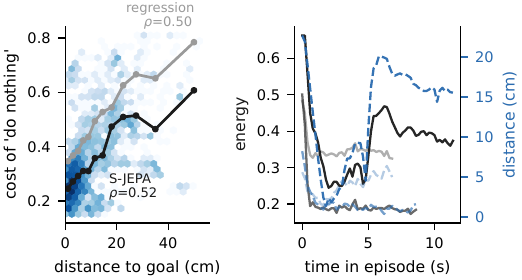}
\caption{The energy reads as distance to the goal. Left: cost of the ``do nothing'' chunk against the true distance of the end effector to the goal pose over 5,561 planning calls of the 100 CALVIN image-goal episodes (density; lines: binned means; $\rho$: Spearman correlation). Right: three completed episodes; the energy and the true distance fall together and flatten once the goal is reached.}
\label{fig:e0}
\end{figure}

\subsection{Energy as distance to the goal (Fig.~\ref{fig:e0})}
\label{app:e0}
During 100 image-goal windows planned with the static-camera predictor (energy at $\tau = 700$, $J = 2$) we log, at every planning call, the energy of the chunk that holds the end effector still (zero pose change, gripper command kept) and the true distance of the end effector to the goal pose. Pooled over 5,561 calls, the Spearman correlation is 0.52 for the energy and 0.50 for the regression predictor's latent distance; binned means rise monotonically with distance (energy: 0.25 at 0--3~cm, 0.29 at 3--6, 0.31 at 6--10, 0.35 at 10--15, 0.48 at 15--25, 0.51 beyond 25~cm). Per window, the median correlation is 0.46. Fig.~\ref{fig:e0} shows the pooled calls and three completed episodes.

\section{DINO-WM benchmarks}
\label{app:dinowm}

\subsection{Data and encoder}
We use the released DINO-WM datasets and models~\cite{dinowm}: PushT (18,685 episodes, frame skip 5), PointMaze (2,000 episodes of 100 frames) and Wall (1,920 episodes of 51 frames). Frames are encoded by the frozen DINOv2 ViT-S/14 at 224~px into 256 tokens of 384 channels, to which a 10-dimensional proprioception embedding and a 10-dimensional action embedding are concatenated.

\subsection{Predictors}
Our predictors keep the released architecture (six layers, 16 heads, MLP width 2048, token width 404) and change only the objective: $x_0$ denoising with the mixed level sampler, horizon $H = 5$ with $h$ drawn per batch, 100k steps of batch 32. The released predictors, trained to regress the next tokens and rolled out step by step, are the baselines. Each predictor trains in 14 hours on one RTX~3090.

\subsection{Protocol}
50 start--goal pairs per environment with the goal five steps ahead on the recorded trajectory; three rounds of model-predictive control, each a cross-entropy search over 300 candidates with 30 elites for 30 rounds (PushT, Wall) or 10 rounds (PointMaze); success thresholds of the released evaluation (PushT: position error below 20 and angle below $\pi/9$; PointMaze: 0.5; Wall: 4.5). Our planner evaluates the energy ($\tau \in \{500, 800\}$, and $\tau = 980$ for the limit rows of PointMaze) with $J$ shared draws per level. The regression rows train the same time-skip architecture to regress the target tokens and plan with the latent distance in one pass.

\subsection{Results}
Table~\ref{tab:app:dinowm} lists success and planning time for the three environments, the rows of Tab.~\ref{tab:maze} of the main text with the two-view PushT rows. A second goal view, the same goal state rendered from a second recorded configuration, raises either planner on PushT.

\begin{table}[h]
\centering
\footnotesize
\setlength{\tabcolsep}{3.5pt}
\begin{tabular}{llccc}
\toprule
environment & planner & goal views & success & time \\
\midrule
PointMaze & DINO-WM rollout & 1 & 1.00 & 145 min \\
 & energy, $J{=}1$ & 1 & 0.96 & 5 min \\
 & energy, $J{=}4$ & 1 & 0.98 & 31 min \\
 & energy, two levels & 1 & 1.00 & 9 min \\
PushT & DINO-WM rollout & 1 & 0.92 & 7.7 h \\
 & DINO-WM rollout & 2 & 0.98 & 7.8 h \\
 & regression, distance & 1 & $0.61^{a}$ & 7 min \\
 & energy, $J{=}4$ & 1 & 0.90 & 50 min \\
 & energy, $J{=}4$ & 2 & 0.96 & 98 min \\
Wall & DINO-WM rollout & 1 & 0.92 & 61 min \\
 & regression, distance & 1 & 0.70 & 8 min \\
 & energy, $J{=}1$ & 1 & 0.90 & 15 min \\
\bottomrule
\end{tabular}
\caption{DINO-WM benchmarks, 50 instances per evaluation. $^{a}$The PushT regression success is the mean over five training seeds, each evaluated on the same 50 start--goal pairs, rounded to two decimal places. The first two energy rows of PointMaze are read at $\tau = 980$, all others at $\tau \in \{500, 800\}$; regression is the time-skip predictor read once. Higher success and lower time are better.}
\label{tab:app:dinowm}
\end{table}

\subsection{Stochastic Wall}
\label{app:stoch}
The environment of Sec.~\ref{sec:stoch} is the Wall environment with two changes. The wall has two door slots, an upper and a lower one, both drawn as doors; at the start of each episode one of them, chosen with probability one half, is open and the other blocks the agent like the wall. The goal state lies one agent diameter beyond the wall at the height midway between the slots, so that the state against the wall opposite the goal shares patches with the goal image. The training set has the size of the original (1,920 episodes of 51 frames) and is collected with the original random policy. Two predictors are trained on it for 100k steps: the DINO-WM predictor with the released code, and S-JEPA with horizon $H = 10$, which covers the path through a slot; the distance row reads the S-JEPA predictor at $\tau = 980$.

\paragraph{Reach probability.} For each of the 50 evaluation instances we draw 300 random chunks, execute each under both door states, and record the fraction of the two executions that end within the success threshold of the goal: $0$, $\tfrac12$ or $1$. Section~\ref{sec:stoch} reports the area under the ROC curve of the negated cost as a detector of the chunks with a non-zero fraction, averaged over instances. Of the 15,000 chunks, 6.1\% reach the goal under one door state.

\paragraph{Crossing level.} For each instance we take the chunk $a$ through the slot nearer the goal and the chunk $a'$ that ends against the wall opposite the goal, and read $E_\tau(a \mid g) - E_\tau(a' \mid g)$ at 20 levels with $J = 16$. The difference is positive at the high levels and negative at the low ones; it changes sign at a median $\tau = 640$ (interquartile range 590--680). This is the behaviour of Eq.~\ref{eq:posterior}: above the crossing the responsibilities equal the priors and the cost is the distance to the mean of the two futures, and below it they select the future at the goal.

\paragraph{Planning.} The protocol is that of the Wall environment (50 instances, three rounds of model-predictive control); the door state is drawn per instance and fixed across planners. The energy planner reaches the goal in 43 of 50 instances, the same predictor read at $\tau = 980$ in 21 and the DINO-WM planner in 19; in 24 of its 31 failures the DINO-WM planner ends against the wall opposite the goal.

\paragraph{Door probability.} The probability that the upper slot is the open one is varied from one half to one. For each probability we train S-JEPA with five training seeds and evaluate both readings of each trained predictor on the same 50 start--goal instances. Every entry in Table~\ref{tab:app:dial} is the mean of the five per-seed success rates, rounded to two decimal places. At probability one the environment is deterministic and the two readings of the predictor plan alike, as Appendix~\ref{app:prop1} states for a single future; the lead of the energy grows as the door becomes uncertain (Table~\ref{tab:app:dial}).

\begin{table}[h]
\centering
\footnotesize
\begin{tabular}{lcccc}
\toprule
probability of the upper slot & 1 & 0.9 & 0.7 & 0.5 \\
\midrule
S-JEPA, distance ($\tau = 980$) & 0.90 & 0.76 & 0.56 & 0.42 \\
S-JEPA, energy & 0.90 & 0.88 & 0.87 & 0.86 \\
\bottomrule
\end{tabular}
\caption{Stochastic Wall as the door becomes uncertain. Every entry in both rows is the mean success rate over five training seeds, each evaluated on the same 50 instances, rounded to two decimal places. Higher is better.}
\label{tab:app:dial}
\end{table}

\section{Navigation with the released NWM}
\label{app:nwm}

\subsection{Model and data}
We use the released Navigation World Model CDiT-B~\cite{nwm} and the 100 evaluation trajectories of its planning benchmark, with no training or fine-tuning. The model is a conditional diffusion transformer over the latent of a frozen image autoencoder, conditioned on the current observation, a relative action (displacement and yaw) and a relative time.

\subsection{Official planner}
The released planner samples 50-step action trajectories with the cross-entropy method, simulates each with the diffusion model, and ranks by the distance between the decoded endpoint and the goal image (LPIPS and the latent distance); with two goal views the ranking sums both. On one RTX~3090 running nothing else it takes 188~s per trajectory with one goal view and 189~s with two.

\subsection{Energy}
We read Eq.~\ref{eq:energy} from the same network: the goal image is encoded, noised at a set of levels, and the denoising error of the released model conditioned on the candidate's action and time is the cost. Settings: $T = 8$ levels in $[50, 750]$ with the $x_0$ weighting, $J = 2$ shared draws, one round of 120 candidates (Appendix~\ref{app:nwmabl}). With several goal views the energies add (Eq.~\ref{eq:product}).

\subsection{Results}
Table~\ref{tab:app:nwm} lists the variants behind Tab.~\ref{tab:nwm} of the main text: the level set and its weighting, and one, two and four goal views. Every energy row plans a trajectory in under 20~s against 188~s for the official planner, and each added view lowers the error of the energy as it does for the official cost. Cut to the three candidates that it can roll out in the time of the one-view energy, the official planner reaches 1.71.

\begin{table}[h]
\centering
\footnotesize
\setlength{\tabcolsep}{3.5pt}
\begin{tabular}{lcccc}
\toprule
planner & goal views & ATE & RPE & s/traj. \\
\midrule
no planning (forward) & -- & 1.856 & 0.519 & -- \\
official, 1 view & 1 & 1.107 & 0.345 & 188 \\
official, 2 views & 2 & 0.922 & 0.326 & 189 \\
official, 3 candidates & 1 & 1.71 & 0.498 & 4.7 \\
energy, uniform weighting & 1 & 1.19 & 0.338 & 4.5 \\
energy, $x_0$ weighting (reported) & 1 & 1.12 & 0.331 & 4.5 \\
energy, $x_0$ weighting, 2 views & 2 & 0.93 & 0.309 & 8.9 \\
energy, $x_0$ weighting, 4 views & 4 & 0.85 & 0.301 & 18 \\
\bottomrule
\end{tabular}
\caption{NWM planning variants, 100 trajectories; times per trajectory on one idle RTX~3090 (first five).}
\label{tab:app:nwm}
\end{table}

\subsection{Search ablations}
\label{app:nwmabl}
With the reported energy and one goal view, one round of the cross-entropy method over 120 candidates gives an absolute trajectory error of 1.12; 480 candidates give 1.10 at four times the cost, three rounds 1.31, and gradient refinement of the chunk 1.52. One round is the setting of every energy row: further rounds and gradient steps move the chunk to regions where the released model was not trained, where the energy is low for reasons unrelated to the goal. With the prior term of Eq.~\ref{eq:map}, a Gaussian fitted to the action chunks of the model's training trajectories, the search stays on that support: three rounds then give 1.08 and gradient refinement gives 1.10 on the same trajectories.

\section{Joint training of the encoder}
\label{app:joint}
\subsection{Recipe}
The encoder (DINOv2 ViT-S/14 on the DINO-WM benchmarks) is trained together with the predictor. Targets are produced by an exponential-moving-average copy of the encoder with no gradient (rate $0.999$ or $0.9995$); the context frames pass through the online encoder, which receives gradients only through the last $k$ frames; a variance hinge $\max(0, 1 - \mathrm{std}_j)$ averaged over the channels $j$ of the online context tokens is added with weight $\gamma$; the encoder's learning rate is $10^{-6}$ to $10^{-5}$ against $5\times10^{-4}$ for the predictor. The standardisation statistics of the targets are recomputed from the EMA encoder.

\subsection{Collapse without the hinge}
\begin{figure}[h]
\centering
\includegraphics[width=\linewidth]{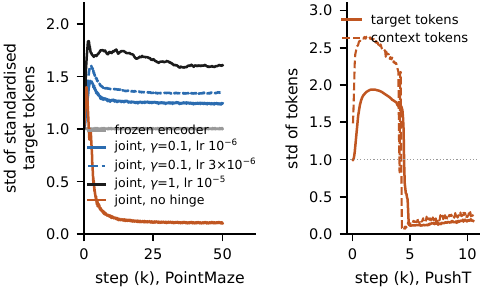}
\caption{The variance hinge keeps the target tokens spread. Left: standard deviation of the standardised target tokens over training for the PointMaze runs of Table~\ref{tab:app:joint}. Right: the PushT run with an EMA target and no hinge; the standard deviation of the target tokens (solid) and of the online context tokens (dashed) falls below $0.25$ within 10k steps while the loss keeps falling.}
\label{fig:app:collapse}
\end{figure}

With the EMA target (rate $0.998$) but no variance term, on PushT, the standard deviation of the standardised target tokens falls from $1.00$ to $0.18$ within 10k steps (minimum $0.11$) and that of the online context tokens from $1.48$ to $0.24$, while the training loss keeps falling: a target that is constant across the patches of a frame is a minimiser of the denoising objective, since the noise is then the deviation of each noised patch from their mean (Fig.~\ref{fig:app:collapse}). On PointMaze the same control ($\gamma = 0$, EMA $0.999$, learning rate $10^{-5}$) falls below $0.25$ within 6k steps and ends at $0.07$ after 50k, with a raw token std of $0.08$ against $1.29$ for the frozen encoder: the patches of a frame have collapsed onto one vector. That vector still depends on the frame. The position probe reaches $R^2$ $0.995$ and the planner completes 49 of 50 instances (Table~\ref{tab:app:joint}, last row), since the maze state is one global quantity that a single vector per frame carries. The hinge keeps the per-patch structure of the tokens, which PushT tests (Table~\ref{tab:app:jointpusht}).

\subsection{Runs and results}
Five PointMaze predictors are trained for 50k steps each: a frozen-encoder reference, three jointly trained encoders ($\gamma = 0.1$, EMA $0.999$, encoder learning rate $10^{-6}$; $\gamma = 0.1$, EMA $0.9995$, $3\times10^{-6}$; $\gamma = 1.0$, EMA $0.999$, $10^{-5}$) and a control without the hinge ($\gamma = 0$, EMA $0.999$, $10^{-5}$). Table~\ref{tab:app:joint} reports, for each, the drift of the standardised target tokens, the spread of the raw encoder tokens within a frame, planning success under the DINO-WM PointMaze protocol (energy at $\tau = 980$, $J = 4$, the setting of Table~\ref{tab:maze}), and a linear probe of the maze position from the mean and standard deviation over tokens of a single frame (ridge regression on 1,800 training frames, $R^2$ on 200 held-out frames; velocities are not decodable from one frame for any encoder).

\begin{table}[h]
\centering
\footnotesize
\setlength{\tabcolsep}{3.5pt}
\begin{tabular}{lcccc}
\toprule
encoder (PointMaze) & target std & token std & success & position $R^2$ \\
\midrule
frozen (reference) & 0.99 & 1.29 & 0.98 & 0.995 \\
joint, $\gamma{=}0.1$, lr $10^{-6}$ & 1.19 & 1.54 & 1.00 & 0.999 \\
joint, $\gamma{=}0.1$, lr $3{\times}10^{-6}$ & 1.34 & 1.66 & 0.96 & 0.998 \\
joint, $\gamma{=}1.0$, lr $10^{-5}$ & 1.60 & 2.04 & 0.82 & 0.974 \\
joint, no hinge, lr $10^{-5}$ & 0.07 & 0.08 & 0.98 & 0.995 \\
\bottomrule
\end{tabular}
\caption{Joint training of the encoder on PointMaze, 50k steps. Target std: standard deviation of the standardised target tokens against the statistics fitted at step 0 (1 means no drift, near 0 means the patches of a frame have collapsed onto one vector). Token std: mean per-channel standard deviation of the encoder tokens within a frame. Success: 50 instances, energy at $\tau = 980$, $J = 4$. Position $R^2$: ridge regression of the maze position from the single-frame tokens, 200 held-out frames.}
\label{tab:app:joint}
\end{table}

The two encoders trained at hinge weight $0.1$ stay close to the frozen one (raw token std $1.54$ and $1.66$ against $1.29$) and plan and decode the position as well as it. At hinge weight $1$ and learning rate $10^{-5}$ the encoder moves further (raw token std $2.04$) while the target tokens stay spread ($1.60$), and success falls to $0.82$ with $R^2$ $0.974$. The hinge holds the target tokens spread at all three learning rates; the learning rate sets how far the encoder drifts from DINOv2, and at $10^{-5}$ the drift costs planning success. The settings of Sec.~\ref{sec:maze} are the two at hinge weight $0.1$. Without the hinge (last row) the tokens of a frame collapse onto one vector that still carries the maze position, as read above.

\subsection{PushT}
PushT depends on the layout of the patches: the pose of the block is not a single global quantity. Table~\ref{tab:app:jointpusht} repeats the comparison there, 100k steps each, with the energy planner of Tab.~\ref{tab:maze}. Without the hinge the planner loses the goal; with it the jointly trained encoder plans as well as the frozen one.

\begin{table}[h]
\centering
\footnotesize
\setlength{\tabcolsep}{3.5pt}
\begin{tabular}{lccc}
\toprule
encoder (PushT) & target std & token std & success \\
\midrule
frozen (reference) & 1.00 & 1.31 & 0.90 \\
joint, $\gamma{=}0.1$, lr $10^{-6}$ & 1.16 & 1.49 & 0.92 \\
joint, $\gamma{=}0.1$, lr $3{\times}10^{-6}$ & 1.29 & 1.60 & 0.90 \\
joint, no hinge, lr $10^{-5}$ & 0.11 & 0.12 & 0.14 \\
\bottomrule
\end{tabular}
\caption{Joint training of the encoder on PushT. Columns as in Table~\ref{tab:app:joint}; success over 50 instances with the energy at $J = 4$.}
\label{tab:app:jointpusht}
\end{table}

\section{Real-robot study}
\label{app:robot}

\begin{figure*}[t]
\centering
\includegraphics[width=0.92\linewidth]{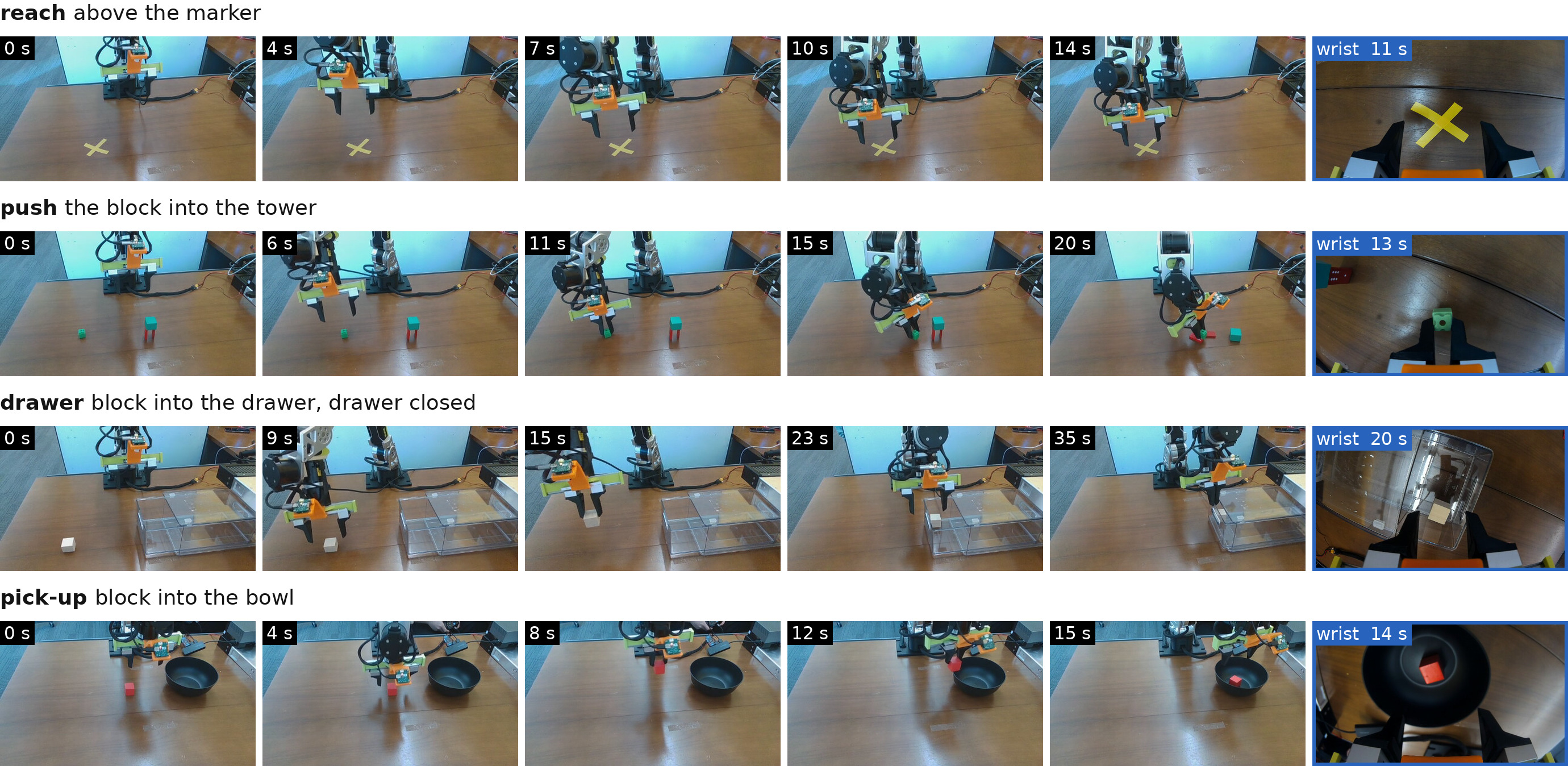}
\caption{Tasks of the same-budget comparison (T1) on the reBot arm: reach, push, drawer and pick-up. Each row is one episode: five frames of the exterior camera and, in the last column, the wrist camera near the contact. Times are seconds from the start of the episode.}
\label{fig:app:robottasks}
\end{figure*}

\begin{figure*}[t]
\centering
\includegraphics[width=0.92\linewidth]{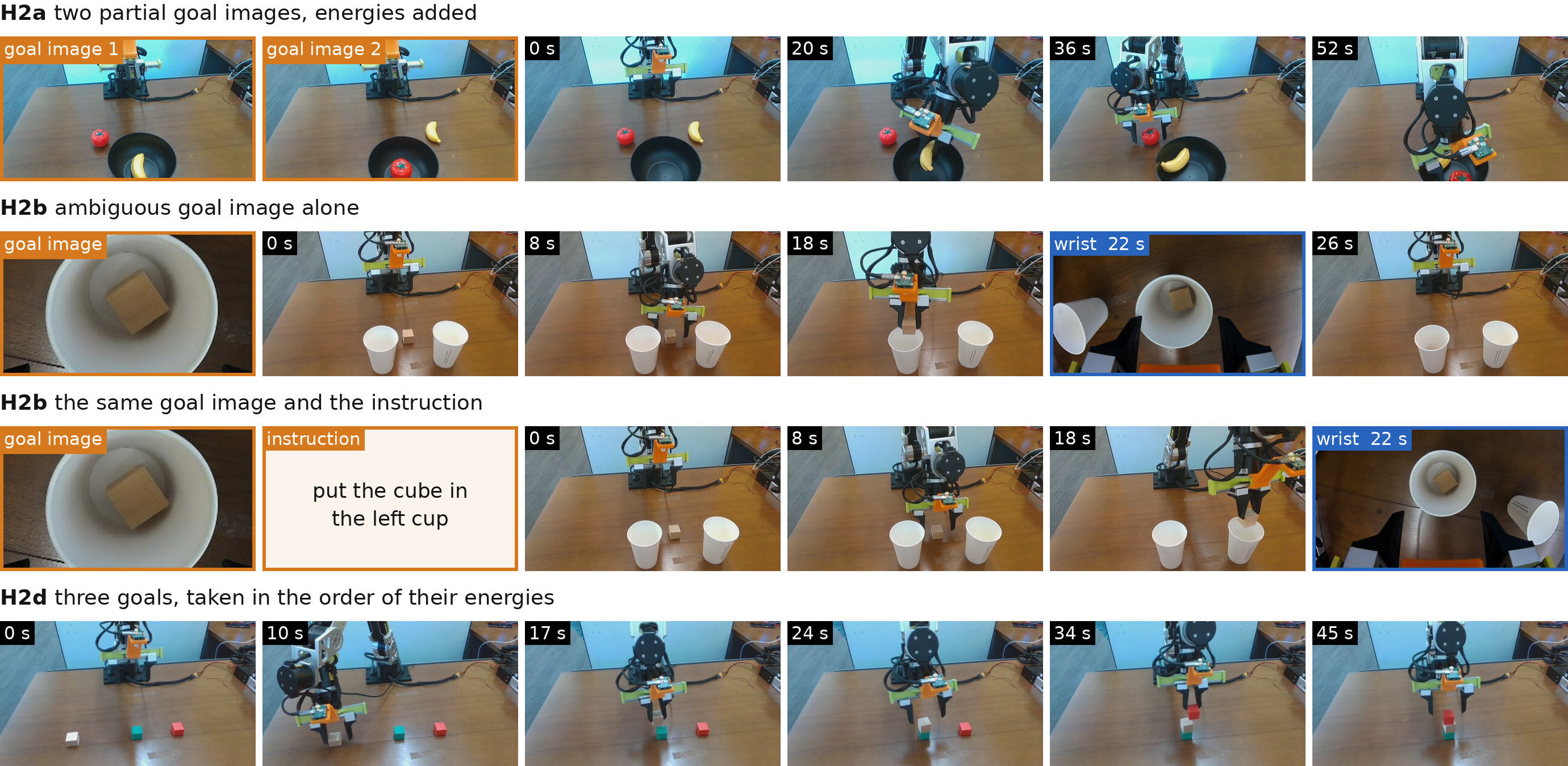}
\caption{Composed goals (H2). H2a: two partial goal images, one with the banana in the bowl and one with the tomato, and an episode planned on the sum of their energies. H2b: a goal image cropped from the wrist camera shows a cube in a cup and leaves open which cup; planned on the image alone the cube goes to the other cup, and with the instruction as a second expert to the left cup of the arm. H2d: three blocks stacked in the order of the energies of their goals. Goal images and the instruction are framed in orange, wrist-camera frames in blue.}
\label{fig:app:roboth2}
\end{figure*}
\begin{table*}[t]
\centering
\footnotesize
\setlength{\tabcolsep}{3pt}
\begin{tabular}{l p{7.0cm} p{3.9cm} p{4.8cm}}
\toprule
id & experiment & trials & what it shows \\
\midrule
H1 & typed instructions, several in a row, no goal image & 3 layouts $\times$ 5 chains $\times$ 4 instr. & language channel, energy as the stop signal \\
H2a & two partial goal images added & 10 (+10 each alone) & product of goals \\
H2b & ambiguous goal image + instruction & 3 goal types $\times$ 10 & product of goals \\
H2d & three goal images, planner orders them by energy & 10 & energy as feasibility \\
H3a--c & stop, retry, missing goal object, moved object & 10 each & energy as monitor; replanning \\
H3d & goal moved by hand at 2 cm/s & 5 & closed-loop rate \\
H5 & goal 2 s away: $0.5$-s vs $2$-s predictor & 10 each & horizon \\
H6 & pick-up and reach with the exterior view, the wrist view, or both & 10 each & views as experts \\
T1 & reach, push, drawer, button, pick-up; ours vs regression & 20 each & same-budget comparison \\
T2 & demonstrations with goals $0.5$--$4$ s ahead & 10 demos, offline & horizon landscape on real data \\
T3 & energy vs true distance from all logs & offline & energy as distance \\
\bottomrule
\end{tabular}
\caption{Real-robot experiments. T1 and H1 are reported in Tab.~\ref{tab:robot} of the main text, the rest in Sec.~\ref{app:robotres}.}
\label{tab:app:robotplan}
\end{table*}

\begin{figure*}[t]
\centering
\includegraphics[width=0.92\linewidth]{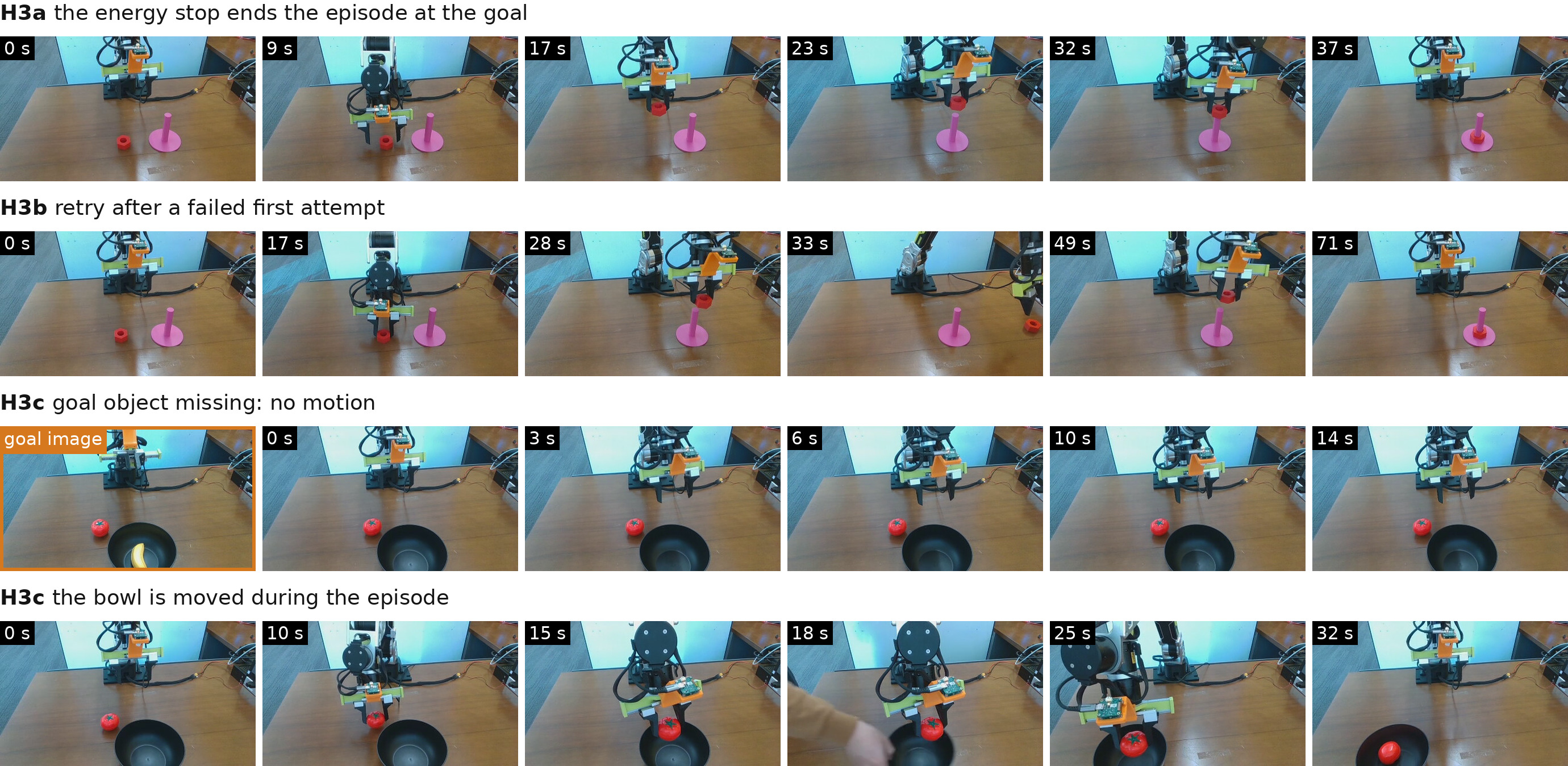}
\caption{The energy as a monitor (H3), exterior camera. H3a: the ring is placed on the peg and the arm retreats and stops. H3b: the first attempt misses and the ring rolls away (33~s); the arm picks it up again and places it. H3c, top: the goal image shows a banana in the bowl and no banana is on the table; the arm stays in place. H3c, bottom: a hand moves the bowl while the arm carries the tomato (18~s), and the arm follows the bowl.}
\label{fig:app:roboth3}
\end{figure*}

\subsection{Hardware and interface}
A single 6-DoF Seeed reBot Arm B601-RS with two RGB cameras, as in the DROID recordings: a fixed one placed to match their exterior view (table in the lower half of the image, base to one side, $30$--$45^\circ$ downward) and one on the wrist that looks down between the fingers; 256-px input per camera; actions are 10-Hz end-effector pose changes with a gripper command, executed by the arm controller with per-step limits on translation and rotation inside a workspace box. The planner runs on a GPU host (one RTX~4090) on the same network and returns, in 1.2~s per call, two actions, the energy of the plan and that of the ``do nothing'' chunk.

\subsection{Model}
One predictor per camera, as in CALVIN, with the architecture of Appendix~\ref{app:predictors} and the energies of the two views added (Eq.~\ref{eq:product}); each is trained on 29.5k DROID episodes (exterior and wrist streams) with the frozen V-JEPA~2 encoder and the language channel on the DROID instructions, then fine-tuned once on 200 trajectories recorded on the reBot arm, the only data of this arm the predictor sees. The fine-tuning is a single run over all five tasks. The regression predictor is trained and fine-tuned on the same data and is the baseline. The horizon experiments H5 and T2 use a predictor trained to $2$~s on the same data.

\subsection{Protocol}
Table~\ref{tab:app:robotplan} lists the experiments; T1 runs 20 trials per task and planner, H1 15 chains of four instructions. Every trial records both camera streams, the proprioceptive state, the executed actions, both energies and the planning time per call, the goal (image, instruction, or set), and the pose of the goal state when the goal image is taken, so that the distance to the goal is known at every step. Success is judged from video by a person who does not know which planner ran; failures are classified as no motion, drift, reached but not completed, collision, or timeout. Ours and the regression planner alternate on the same initial states and goals.

\subsection{Results}
\label{app:robotres}
T1 and H1 are reported in Tab.~\ref{tab:robot} of the main text, and Fig.~\ref{fig:app:robottasks} shows an episode of four of the tasks. Per task, the energy succeeds in 15 to 19 of 20 trials from a goal image against 9 to 14 for the latent distance, and in 18 to 20 from the instruction alone; 11 of the 15 chains are completed end to end, 53 of the 60 instructions; the energy stop agrees with the judge in 87 of the 100 image trials and in 92 of the 100 text trials.

\paragraph{Composed goals (H2).} Table~\ref{tab:app:robotH} lists the trials and Fig.~\ref{fig:app:roboth2} shows an episode of each. Two partial goal images, each showing one of two objects in place, complete the two-object goal in 9 of 10 trials when their energies add, against 4 and 5 for either image alone. A goal image that leaves the task ambiguous completes 12 of 30 trials alone and 26 of 30 with the instruction as a second expert. Given three goal images, their energies order the goals as they can be reached in 9 of 10 scenes.

\paragraph{The energy as a monitor (H3).} Figure~\ref{fig:app:roboth3} illustrates stopping, retrying, an unavailable goal object, and a displaced bowl. The energy stop ends the episode at the goal in 9 of 10 trials. After a failed first attempt the planner retries and completes 8 of 10. When the object shown in the goal image is absent from the scene, the energy of every candidate stays above the stop level in 10 of 10 trials and the arm does not move. When the object is moved during the episode the planner completes 8 of 10, and it follows a goal moved by hand at 2~cm/s in 4 of 5.

\paragraph{Horizon (H5, T2).} With the goal $2$~s away, the predictor trained to $2$~s completes 9 of 10 trials and the predictor trained to $0.5$~s completes 3. On 10 recorded demonstrations the rank of the demonstrated action among perturbed ones is 0.01, 0.38, 0.47 and 0.51 for goals $0.5$, $1$, $2$ and $4$~s ahead with the $0.5$-s predictor, and 0.02, 0.03, 0.05 and 0.31 with the $2$-s predictor, the curve of Fig.~\ref{fig:horizon} on the real arm.

\begin{figure}[!t]
\centering
\includegraphics[width=\linewidth]{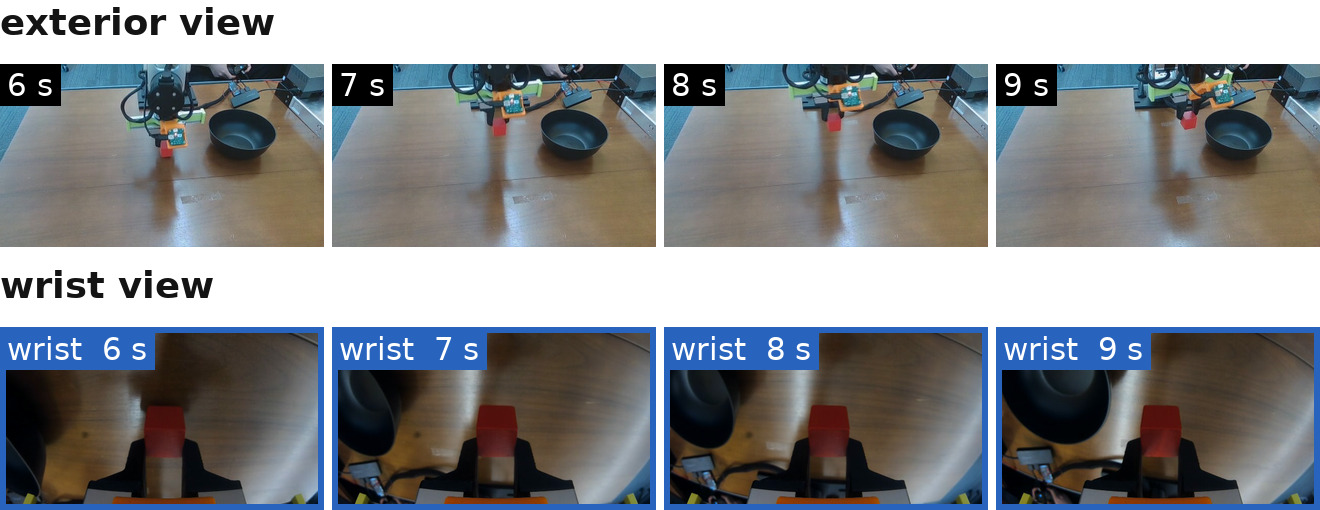}
\caption{The two views during a grasp (H6). In the exterior view the gripper covers the block as the fingers close; the wrist view holds both fingertips and the block in one frame.}
\label{fig:app:robotviews}
\end{figure}

\paragraph{Views (H6).} With the exterior view alone the planner completes 3 of 10 pick-ups, with the wrist view alone 6, and with both 9; reaching completes 9 of 10 with the exterior view alone. The exterior view loses the object behind the gripper as the fingers close, and the wrist view holds both fingertips and the object in one frame (Fig.~\ref{fig:app:robotviews}): the grasp is decided by the view that resolves the contact, as in CALVIN, where the static camera alone completes 3\% of the grasp windows of scene $D$.

\paragraph{Energy and distance (T3).} Over all logged planning calls the Spearman correlation between the energy of the ``do nothing'' chunk and the distance of the end effector to the goal pose is 0.61.

\begin{table}[h]
\centering
\footnotesize
\setlength{\tabcolsep}{4pt}
\begin{tabular}{llc}
\toprule
id & condition & completed \\
\midrule
H2a & first partial goal image alone & 4 / 10 \\
 & second partial goal image alone & 5 / 10 \\
 & both, energies added & 9 / 10 \\
H2b & ambiguous goal image alone & 12 / 30 \\
 & goal image and instruction & 26 / 30 \\
H2d & energy order matches reachable order & 9 / 10 \\
H3a & energy stop at the goal & 9 / 10 \\
H3b & retry after a failed attempt & 8 / 10 \\
H3c & goal object missing: no motion & 10 / 10 \\
 & object moved during the episode & 8 / 10 \\
H3d & goal moved by hand at 2 cm/s & 4 / 5 \\
H5 & goal $2$~s away, $0.5$-s predictor & 3 / 10 \\
 & goal $2$~s away, $2$-s predictor & 9 / 10 \\
H6 & pick-up, exterior view alone & 3 / 10 \\
 & pick-up, wrist view alone & 6 / 10 \\
 & pick-up, both views & 9 / 10 \\
 & reach, exterior view alone & 9 / 10 \\
\bottomrule
\end{tabular}
\caption{Real-robot experiments H2, H3, H5 and H6. Ids as in Table~\ref{tab:app:robotplan}. Higher is better.}
\label{tab:app:robotH}
\end{table}

\section{Limitations}
\label{app:limitations}
The energy ranks goals by the likelihood of their representation under the predictor, so the representation bounds its resolution. A grasp is planned only when one of the views resolves the contact between the fingers and the object: with a frozen encoder at 256~px a fixed camera places the two states a few patches apart, and in CALVIN and on the real arm it is the camera on the gripper that separates them; the horizon result shows how far the cost sees and says nothing about how finely. For the composition of Sec.~\ref{sec:composition}, conditional independence gives a bound on the joint negative log-likelihood by summing the continuous objectives (Appendix~\ref{app:composition}); the reduction to a squared distance to the combined goal holds in the high-noise limit. The CALVIN numbers come from a planner with a frozen encoder and no policy, and do not compare with language-conditioned policies trained on the same annotations. Joint training of the encoder is shown on PointMaze and PushT, with a DINOv2 encoder. The stochastic environment of Sec.~\ref{sec:stoch} has two futures per chunk and one hidden binary variable. The two-level, one-draw reading of the truncated bound is compared with the full quadrature on CALVIN only.

\end{document}